\documentclass{article}

\usepackage{arxiv}

\usepackage[utf8]{inputenc} % allow utf-8 input
\usepackage[T1]{fontenc}    % use 8-bit T1 fonts
\usepackage{hyperref}       % hyperlinks
\usepackage{url}            % simple URL typesetting
\usepackage{amsfonts}       % blackboard math symbols
\usepackage{nicefrac}       % compact symbols for 1/2, etc.
\usepackage{microtype}      % microtypography
\usepackage{lipsum}
\usepackage{graphicx}
\usepackage{subcaption} 
\usepackage{listings}
\usepackage{fancyhdr}
\usepackage{graphicx}
\usepackage{algorithm}
\usepackage[utf8]{inputenc}
\usepackage[english]{babel}
\usepackage{booktabs}          
\usepackage{amsmath}        
\usepackage[T1]{fontenc}
\usepackage{amssymb,amsmath,hyperref}
\usepackage{graphicx}
\usepackage{transparent}
\usepackage{tikz}
\usepackage{multicol}
\usepackage{amsmath} 
\usepackage{array}
\usepackage{mathtools}
\usepackage{diagbox}
\usepackage{amsfonts}
\usepackage{pgfplots}
\usepackage{hyperref}
\usepackage[dvipsnames]{xcolor}
\usepackage{soul}
\usepackage[normalem]{ulem}
\usepackage{cancel}
\usepackage{amsmath} 
\usepackage{amsmath}  
\usepackage{algorithm}      
\usepackage{algpseudocode}  
\usepackage{float} 
\graphicspath{ {./images/} }

\title{Efficient Score Pre-computation for Diffusion Models via Cross-Matrix Krylov Projection}

\author{
 Kaikwan Lau \\
  David R. Cheriton School of Computer Science\\
  University of Waterloo\\
  Waterloo, ON \\
  \texttt{kai-kwan.lau@uwaterloo.ca} \\
   \And
 Andrew S. Na \\
  David R. Cheriton School of Computer Science\\
  University of Waterloo\\
  Waterloo, ON \\
  \texttt{andrew.na@uwaterloo.ca} \\
  \And
 Justin W.L. Wan \\
  David R. Cheriton School of Computer Science\\
  University of Waterloo\\
  Waterloo, ON \\
  \texttt{justin.wan@uwaterloo.ca} \\
}

\begin{document}
\maketitle

\begin{abstract}

This paper presents a novel framework to accelerate score-based diffusion models. It first converts the standard stable diffusion model into the Fokker-Planck formulation which results in solving large linear systems for each image. For training involving many images, it can lead to a high computational cost. The core innovation is a cross-matrix Krylov projection method that exploits mathematical similarities between matrices, using a shared subspace built from ``seed" matrices to rapidly solve for subsequent ``target" matrices. Our experiments show that this technique achieves a 15.8\% to 43.7\% time reduction over standard sparse solvers. Additionally, we compare our method against DDPM baselines in denoising tasks, showing a speedup of up to 115$\times$. Furthermore, under a fixed computational budget, our model is able to produce high-quality images while DDPM fails to generate recognizable content, illustrating our approach is a practical method for efficient generation in resource-limited settings.
\end{abstract}

% keywords can be removed

\section{Introduction}

Score-based diffusion models have emerged as a powerful paradigm for generative modeling \cite{song2019generative}, achieving state-of-the-art results in image synthesis \cite{ho2020denoising, song2020score}. These models learn to reverse a forward diffusion process by estimating the score function (gradient of log-density) at each timestep. However, this capability comes at a significant computational cost: training requires thousands of epochs on extensive datasets, and inference demands hundreds of denoising steps. These requirements result in high energy consumption and substantial computational expense.

Recent efforts to address these inefficiencies have primarily focused on two directions \cite{luo2023latent}. The first approach accelerates sampling through deterministic samplers such as DDIM \cite{song2020denoising} or improved ODE solvers \cite{lu2022c}, while the second improves training techniques \cite{song2020improved}. Despite these advances, the training bottleneck remains largely unaddressed. While some progress has been made in finite-difference score matching \cite{pang2020efficient}, however, limited work has focused on improving the fundamental score learning process itself. This computational challenge motivates our comprehensive framework that accelerates score-based diffusion models through two synergistic innovations:

\textbf{Pre-computed Score Embedding Framework:} Rather than learning scores from scratch, we build upon the work of Na et al. \cite{na2024efficient}, who pre-computed accurate scores for single-image denoising by numerically solving the log-density Fokker-Planck equation before training. These scores are then embedded into images via transport equations, providing the neural network with high-quality initialization and enabling SSIM-based \cite{wang2004image} early stopping. In this work, we extend their method to multi-image generation, achieving significant efficiency improvements in computational efficiency.

\textbf{Timestep-wise Krylov Projection for Fokker-Planck Solving:} We develop a novel timestep-wise Krylov projection method \cite{chan1999galerkin} that exploits structural similarities between matrices across related images. By first solving a ``seed" image completely to build timestep-specific Krylov subspaces, and then applying these subsequent ``target" images, we are able to achieve significant time reduction in score computation while maintaining numerical accuracy.

Our main contributions are:
\begin{itemize}
\item We introduce a cross-matrix Krylov subspace methodology enabling dramatic speedup for multi-image pre-computation.
\item We achieve efficient projection-based acceleration without quality degradation.
\item We demonstrate that fewer computational steps can achieve equivalent image quality.
\item We present a complete framework optimized for practical multi-image generation scenarios.
\end{itemize}
Our approach enables efficient score-based model training on large-scale datasets while maintaining generation quality. 
%To understand how our method achieves these improvements, we first establish the mathematical foundation underlying score-based diffusion models.

\section{Problem Formulation}
\label{finite difference method}

This section presents the core mathematical framework that motivates our acceleration approach. We begin with the standard formulation of score-based diffusion models, then derive the Fokker-Planck equations that form the computational basis of our method, and finally establish the discretization that leads to the linear systems we seek to accelerate.

\subsection{Score-Based Diffusion Models}

Score-based diffusion models are a class of generative models, such as image generation, that operate by learning to reverse a predefined process of gradually adding noise to data \cite{song2020score}. The core idea involves a forward process, where an image is systematically corrupted over hundreds of steps until it becomes pure random noise, and a reverse process, where a deep neural network learns to undo this corruption one step at a time \cite{song2020score}. This network is trained to estimate the score function, a guiding gradient that directs the denoising at each stage. The model's significant processing time is a direct consequence of this reverse process, which requires an iterative sequence of hundreds of steps to transform noise into an image. Each step in this chain demands a computationally expensive evaluation of the neural network, compounding the total generation time \cite{ho2020denoising}, especially for multi-image training. The foundation of score-based diffusion models \cite{song2020score} rests on stochastic differential equations (SDE). Consider the forward stochastic differential equation:
\begin{equation}
dx = f(x,t)dt + g(t)dW_t,
\label{eq:forward_sde}
\end{equation}
where $t \in [0,t_{\text{end}}]$, $t_{\text{end}}$ is the end time, $x \in \mathbb{R}^D$, $f: \mathbb{R}^D \times [0,t_{\text{end}}] \rightarrow \mathbb{R}^D$ is the drift function, $g: [0,t_\text{end}] \rightarrow \mathbb{R}$ is the diffusion coefficient, and $W_t$ is a $D$-dimensional Wiener process. The generative process requires reversing this forward diffusion, which is achieved through the corresponding reverse SDE \cite{anderson1982reverse}:
\begin{equation}
dx = [f(x,t) - g^2(t)\nabla_x \log p(x,t)]dt + g(t)d\bar{W}_t,
\label{eq:reverse_sde}
\end{equation}
where $p(x,t)$ is the probability density at time $t$, and $\bar{W}_t$ is a reverse-time Wiener process. Both forward and reverse processes involves Wiener process ($W_t$ and $\bar{W}_t$), resulting in a long-time calculation, as the numerical solver must simulate this random path over hundreds or thousands of discrete time steps \cite{kloden1992numerical} in order to achieve a highly accurate solution. 

\subsection{Fokker-Planck Equation Derivation}

Instead of solving the SDE in Eq.\eqref{eq:forward_sde}, which involves a prohibitive computational cost, the evolution of the probability density $p(x,t)$ under the forward SDE \eqref{eq:forward_sde} is satisfied by the Fokker-Planck equation, which does not involve the Wiener process. The equation is described by Øksendal \cite{oksendal2013stochastic}:
\begin{equation}
%        p_t(x,t) =  - \sum_{i=1}^{D} \frac{\partial [f_i(x,t) p(x,t)]}{\partial x_i}   
\frac{\partial p}{\partial t} =  - \sum_{i=1}^{D} \frac{\partial [f_i(x,t) p(x,t)]}{\partial x_i}   
         + \sum_{i,j=1}^{D} \frac{\partial^2 [\textbf{G}_{ij}(x,t) p(x,t)]}{\partial x_i \partial x_j} ,
    \label{Øksendal}
\end{equation} 
% \begin{equation}
%         p_t(x,t) =  - \sum_{i=1}^{D} \frac{\partial}{\partial x_i} [f_i(x,t) p(x,t)]  
%          + \sum_{i,j=1}^{D} \frac{\partial^2}{\partial x_i \partial x_j} [\textbf{D}_{ij}(x,t) p(x,t)],
%     \label{Øksendal}
% \end{equation} 
%%%%%%%%%%%%%%%%%%%%%%%%%%%%%%%%%%%%%%%%%%%%
% \begin{equation}
%     \begin{aligned}
%         \frac{\partial}{\partial t} p(x,t) =&  - \sum_{i=1}^{D} \frac{\partial}{\partial x_i} [f_i(x,t) p(x,t)] \\  
%          &+ \sum_{i=1}^{D} \sum_{j=1}^{D} \frac{\partial^2}{\partial x_i \partial x_j} [\textbf{D}_{ij}(x,t) p(x,t)],
%     \end{aligned}
%     \label{Øksendal}
% \end{equation} 
where the tensor $\textbf{G} := \frac{1}{2} g(t) g(t)^T = \frac{1}{2} g^2(t)$, $x=(x_1,x_2,\dots,x_D)$, and $f(x,t) = (f_1, f_2, f_3, \dots, f_D) $. We substitute $\textbf{G}$ into Eq.\eqref{Øksendal} and obtain
\begin{equation}
\frac{\partial p}{\partial t} = - \sum_{i=1}^{D} \frac{\partial [f_i(x,t) p(x,t)]}{\partial x_i}  + \frac{g^2(t)}{2} \sum_{i,j=1}^{D}  \frac{\partial^2 p(x,t)}{\partial x_i \partial x_j} .
    \label{Øksendal gt}
\end{equation}
Note $p(x,t)$ can be written as $\prod_{k=1}^{D} p_k(x_k(t))$, as $\{ x_1,x_2,\dots,x_D \}$ are independent and identically distributed random variables. The right-hand side of Eq.(\ref{Øksendal gt}) can be written as

\begin{eqnarray}
      - \sum_{i=1}^{D} \frac{\partial}{\partial x_i} [f_i(x, t) p(x, t)] + \frac{g(t)^2}{2}  [p(x, t) \nabla^2 \log p(x, t)] \nonumber 
      +\frac{g(t)^2}{2} p(x, t) \langle \nabla_x \log p(x, t), \nabla_x \log p(x, t) \rangle.
%\end{split}
%\end{equation}
\end{eqnarray}
Letting $m = \log p$, the log-density $m(x,t)$ satisfies the non-linear partial differential equations (PDE):
\begin{equation}
\frac{\partial m}{\partial t} = - \text{Div} f - \langle f, \nabla_x m \rangle + \frac{1}{2} g^2 \nabla^2 m + \frac{1 }{2}g^2 \langle \nabla_x m, \nabla_x m \rangle.
    \label{eq:log_fp}
\end{equation}
We refer to the Appendix in the Supplementary Materials for the detailed derivation. 
% The Eq.\eqref{eq:log_fp} provides the theoretical foundation for the log-density $m(x,t)$, but it is extremely difficult to look for an exact solution. Instead, we implement numerical methods to solve this PDE efficiently.

\subsection{Finite Difference Discretization}

The Eq.\eqref{eq:log_fp} provides the theoretical foundation for the log-density $m(x,t)$, but it is difficult to look for an exact solution from the non-linear PDE. To obtain the score $\nabla_x m$, we implement numerical methods by discretizing the Eq.\eqref{eq:log_fp} using a standard five-point stencil. The dimension of the Eq.\eqref{eq:log_fp} is two, as we are considering 2-dimensional images with width $W$ and height $H$. Let $t_n$ be the timestep, for all $n \in \mathbb{N}$, $t_n=(n-1)(\Delta t) $, where $\Delta t = \frac{1}{T}$ is the temporal step size, $T$ is the total timesteps, $\Delta x = \frac{1}{H} $, and $\Delta y = \frac{1}{W}$. Let $x_n:=(n-1)(\Delta x)$ and $y_n:=(n-1)(\Delta y)$ be the discretized images over the pixels of each channel. We use zero padding at the borders of the image as the boundary condition. Let $\Delta x = \Delta y =h$ be the spatial grid spacing, 
and let $m_{i,j}^n$, $f^n_{i,j}$, and $g^n$ be approximations to $m(x_i,y_j,t_n)$, $f(x_i,y_j,t_n)$, and $g(t_n)$, respectively. 
We discretize the derivatives of the log-density distribution as follows:
\begin{align}
%        m_{i,j}^n :=& m(x_i,y_j,t_n), \\
%        f^n_{i,j} :=& f(x_i,y_j,t_n), \\
%        g^n :=& g(t_n), \\
        m_t \approx& \frac{m_{i,j}^n - m_{i,j}^{n-1}}{\Delta t} ,\\
    \nabla_x m \approx& \left( \frac{m_{i+1,j}^n - m_{i-1,j}^n}{2h}, \frac{m_{i,j+1}^n - m_{i,j-1}^n}{2h} \right), \\
    \nabla^2 m \approx& \frac{m_{i+1,j}^n + m_{i,j+1}^n - 4m_{i,j}^n + m_{i-1,j}^n + m_{i,j-1}^n}{h^2}, \\
        \text{Div} f^n \approx& \frac{(f^n_{i+1,j})_x - (f^n_{i-1,j})_x}{2 h} + \frac{(f^n_{i,j+1})_y - (f^n_{i,j-1})_y}{2h},
\end{align}
where $f^n_{i,j}=((f^n_{i,j})_x,(f^n_{i,j})_y)$. The main computational challenge arises from the non-linear term $\langle \nabla_x m^n, \nabla_x m^n \rangle$. To avoid this, we apply an iterative method to do the approximation. The non-linear term $\langle \nabla_x m^n, \nabla_x m^n \rangle$ is linearized using a semi-explicit approach:
\begin{equation}
\langle \nabla_x m^n, \nabla_x m^n \rangle \approx \langle \nabla_x\tilde{m}^{n} , \nabla_x m^n \rangle,
\end{equation}
where $\nabla_x \tilde{m}^{n}$ is the approximation from the previous iteration, that is $\nabla_x \tilde{m}^{n,(k)}=\nabla_x m^{n,(k-1)}$, where $k \in \mathbb{N}$ refers to $k$-th iteration. This leads to the discretized system:

\begin{equation}
\begin{split}
        C_{\text{diag}} m^{n,(k)}_{i,j} 
          + C_{\text{east}} m^{n,(k)}_{i+1,j} 
          + C_{\text{north}} m^{n,(k)}_{i,j+1}  
          + C_{\text{west}} m^{n,(k)}_{i-1,j} 
         + C_{\text{south}} m^{n,(k)}_{i,j-1} 
          = b^{n,(k)}_{i,j},
\end{split}
\label{discretized system}
\end{equation}
where
\begin{align}
C_{\text{diag}} &= \frac{1}{\Delta t} + \frac{2(g^n)^2}{h^2}, \\
C_{\text{north}} &= -\frac{(g^n)^2}{2h^2} + \frac{(f^n_{i,j})_y}{2h} - \frac{(g^n)^2}{8h^2}(\tilde{m}_{i,j+1}^{n,(k)} - \tilde{m}_{i,j-1}^{n,(k)}), \\
C_{\text{south}} &= -\frac{(g^n)^2}{2h^2} - \frac{(f^n_{i,j})_y}{2h} + \frac{(g^n)^2}{8h^2}(\tilde{m}_{i,j+1}^{n,(k)} - \tilde{m}_{i,j-1}^{n,(k)}), \\
C_{\text{east}} &= -\frac{(g^n)^2}{2h^2} + \frac{(f^n_{i,j})_x}{2h} - \frac{(g^n)^2}{8h^2}(\tilde{m}^{n,(k)}_{i+1,j} - \tilde{m}^{n,(k)}_{i-1,j}), \\
C_{\text{west}} &= -\frac{(g^n)^2}{2h^2} - \frac{(f^n_{i,j})_x}{2h} + \frac{(g^n)^2}{8h^2}(\tilde{m}^{n,(k)}_{i+1,j} - \tilde{m}^{n,(k)}_{i-1,j}), \\
b^{n,(k)}_{i,j} &= \frac{m^{n-1,(k)}_{i,j}}{\Delta t} - \frac{(f^n_{i+1,j} - f^n_{i-1,j})_x + (f^n_{i,j+1} - f^n_{i,j-1})_y}{2 h}.
\end{align}
(Refer to the Appendix for the detailed derivation.) The discretized equation in Eq.\eqref{discretized system} can be written in matrix form:
\begin{equation}
A^{n,(k)} m^{n,(k)} = b^{n,(k)},
\label{eq:linear_system}
\end{equation}
where the coefficient matrix $A^{n,(k)} \in \mathbb{R}^{HW \times HW}$ has a block tridiagonal structure:
\begin{equation}
A^{n,(k)} = \begin{bmatrix}
M & U & 0 & \cdots & 0 \\
L & M & U & \cdots & 0 \\
0 & L & \ddots & \ddots & \vdots \\
\vdots & \ddots & \ddots & M & U \\
0 & \cdots & 0 & L & M
\end{bmatrix}.
\end{equation}
The diagonal blocks $M \in \mathbb{R}^{H \times H}$ are tridiagonal matrices:
\begin{equation}
M = \begin{bmatrix}
C_{\text{diag}} & C_{\text{north}} & 0 & \cdots \\
C_{\text{south}} & C_{\text{diag}} & C_{\text{north}} & \cdots \\
0 & C_{\text{south}} & \ddots & \ddots \\
\vdots & \ddots & \ddots & \ddots
\end{bmatrix}.
\end{equation}
% \begin{align}
% C_{\text{diag}} &= \frac{1}{\Delta t} + \frac{2(g^n)^2}{h^2}, \\
% C_{\text{north}} &= \frac{1}{2h}\left(-\frac{(g^n)^2}{h} + (f^n_{i,j})_y - \frac{(g^n)^2}{4h}(\tilde{m}_{i,j+1}^{n,(k)} - \tilde{m}_{i,j-1}^{n,(k)})\right), \\
% C_{\text{south}} &= \frac{1}{2h}\left(-\frac{(g^n)^2}{h} - (f^n_{i,j})_y + \frac{(g^n)^2}{4h}(\tilde{m}_{i,j+1}^{n,(k)} - \tilde{m}_{i,j-1}^{n,(k)})\right), \\
% C_{\text{east}} &= \frac{1}{2h}\left(-\frac{(g^n)^2}{h} + (f^n_{i,j})_x - \frac{(g^n)^2}{4h}(\tilde{m}^{n,(k)}_{i+1,j} - \tilde{m}^{n,(k)}_{i-1,j})\right), \\
% C_{\text{west}} &= \frac{1}{2h}\left(-\frac{(g^n)^2}{h} - (f^n_{i,j})_x + \frac{(g^n)^2}{4h}(\tilde{m}^{n,(k)}_{i+1,j} - \tilde{m}^{n,(k)}_{i-1,j})\right), \\
% b^{n,(k)}_{i,j} &= \frac{m^{n-1,(k)}_{i,j}}{\Delta t} - \frac{(f^n_{i+1,j} - f^n_{i-1,j})_x + (f^n_{i,j+1} - f^n_{i,j-1})_y}{2 h}.
% \end{align}
The diagonal blocks $L,U \in \mathbb{R}^{H \times H}$ are defined by:
\begin{align}
    L &= \text{diag}(C_{\text{West}}, C_{\text{West}}, \dots, C_{\text{West}})_{H \times H} ,\\
    U &= \text{diag}(C_{\text{East}}, C_{\text{East}}, \dots, C_{\text{East}})_{H \times H} .
\end{align}
We define the vectors $m^{n,(k)}$ and $b^{n,(k)}$ as follows:
\begin{align}
m^{n,(k)} &= \begin{pmatrix} m^{n,(k)}_{1,1}  m^{n,(k)}_{1,2}  \dots  m^{n,(k)}_{1,H} m^{n,(k)}_{2,1} \dots  m^{n,(k)}_{W,H} \end{pmatrix}^T, \\
b^{n,(k)} &= \begin{pmatrix} b^{n,(k)}_{1,1}  b^{n,(k)}_{1,2} \dots b^{n,(k)}_{1,H} b^{n,(k)}_{2,1} \dots  b^{n,(k)}_{W,H} \end{pmatrix}^T.
\end{align}

After solving this system of linear equations using an iterative method until convergence at a fixed error tolerance with $K$-iteration, the score is obtained by central difference:
\begin{equation}
    \nabla_x m^n = \left( \frac{m^{n,(K)}_{i+1,j}-m^{n,(K)}_{i-1,j}}{2},\frac{m^{n,(K)}_{i,j+1}-m^{n,(K)}_{i,j-1}}{2} \right).
\end{equation}

To summarize, by solving the linear system \eqref{eq:linear_system}, we obtain the score $\nabla_x m^n$ for all timesteps $n$ for an image. To solve the systems efficiently, the LU factorization is applied due to the block structure of the sparse matrix \cite{gilbert1988sparse}. 
% Diffusion models training usually requires hundred or thousands of images for training. In the above, we look for one image that requires solving for $K$ iterations and $T$ timesteps. If we consider $N$ images, the computational cost would be $N$ times larger, resulting in an inefficient bottleneck, especially for a large image (128×128), which requires to solve a 16384×16384 system of linear equations. Solving system of linear equations becomes computationally expensive as the matrix size scales quadratically with image resolution. To reduce the solution time, we exploit the structural similarities between images.

\section{Timestep-wise Krylov Projection Method for Pre-computing Score}

Diffusion models training usually requires hundreds or thousands of images for training. In the above, we look for one image that requires solving for $K$ iterations and $T$ timesteps. Suppose there are $N$ images, we would have $N$ different matrices $A^{n}_1, A^{n}_2, ..., A^{n}_N$. The computational cost would be $N$ times larger, resulting in an inefficient bottleneck.
%especially for a large image (128×128), which requires solving a 16384×16384 system of linear equations. 
Solving systems of linear equations becomes computationally expensive as the matrix size scales quadratically with image resolution. 

To improve the efficiency in solving different systems of linear equations in \eqref{eq:linear_system}, our idea is to explore the similar information of the matrices $A^{n}_i$. More precisely, instead of solving linear systems independently by LU factorization, we will employ a Krylov subspace iterative method. Importantly, we devise a projection method such that we can produce accurate initial guesses for other linear systems after only one linear system is solved by taking advantage of the similarity of the matrices.
The insight behind our approach lies in the observation that images from real-world datasets are often similar. 
%When training a diffusion model, the images used are often highly similar. 
For instance, faces in CelebA exhibit common facial structures, and vehicles in CIFAR-10 share geometric patterns. We observe that the matrices $A^{n}_1, A^{n}_2, ..., A^{n}_N$ across different images share similar structures. This similarity motivates a Krylov subspace approach where solutions from previous images can accelerate convergence \cite{chan1999galerkin} for subsequent images.
To illustrate this concept intuitively, imagine solving a series of very similar puzzles. Instead of starting each puzzle from scratch, you could learn a general strategy from the first puzzle and apply it to subsequent puzzles, solving them much faster. 

Our Cross-matrix Krylov projection operates on this principle: it learns the underlying mathematical structure from a ``seed" system matrix $CA^n_1$ and uses that knowledge to accelerate the processing of subsequent, similar matrices $ A^n_2,A^n_3,..., A^n_N $. We consider Krylov subspaces because they offer a highly efficient strategy for finding an excellent approximate solution to a large linear system \cite{saad2003iterative}. The key idea is dimension reduction: instead of solving the linear of its original size, the method projects it into a much smaller, manageable Krylov subspace. 
%This approach is effective because the subspace is not random. 
For a matrix $A \in \mathbb{R}^{HW \times HW}$ and vector $r_0 \in \mathbb{R}^{HW}$, the $a$-th Krylov subspace is given by \cite{saad2003iterative}:
\begin{equation}
\mathcal{K}_a(A, r_0) = \text{span}\{r_0, Ar_0, A^2r_0, \ldots, A^{a-1}r_0\}.
\end{equation}
The subspace captures the most relevant dynamics of the system. Consequently, the simplified problem solved within this information-rich subspace yields a solution that is close to the true one, but with a fraction of the computational cost \cite{saad2003iterative}.

\subsection*{Timestep-wise Projection Algorithm}

Unlike traditional Krylov methods that use global subspaces, our approach stores a separate Krylov matrix $V_a^n \in \mathbb{R}^{D^2 \times a}$ of orthonormal basis vectors that span the Krylov subspace $\mathcal{K}_a(A^n, r_0)$ for each timestep $t_n$. 
% Given the initial set of linearly independent vectors that form a basis for a Krylov subspace:
% \begin{equation}
% S = \{ \mathbf{r}_0, A\mathbf{r}_0, A^2\mathbf{r}_0, \dots, A^{a-1}\mathbf{r}_0 \}.    
% \end{equation}
% We apply the Gram-Schmidt process \cite{strang2022introduction} to produce an orthonormal set $\{ \mathbf{q}_1, \mathbf{q}_2, \dots, \mathbf{q}_a \}$. We construct an intermediate set of orthogonal vectors $\{ \mathbf{u}_1, \mathbf{u}_2, \dots, \mathbf{u}_a \}$. Let $\mathbf{u}_1 = \mathbf{r}_0 $ and 
% \begin{equation}
%     \mathbf{u}_e = A^{e-1}\mathbf{r}_0 - \sum_{j=1}^{e-1} \frac{(A^{e-1}\mathbf{r}_0) \cdot \mathbf{u}_j}{\|\mathbf{u}_j\|^2} \mathbf{u}_j, \quad \forall e \in \{ 1,2,...,a\}.
% \end{equation}
% Next, we normalize each orthogonal vector $\mathbf{u}_e$ to obtain the final orthonormal vectors $\mathbf{q}_e$ by $\mathbf{q}_e = \frac{\mathbf{u}_e}{\|\mathbf{u}_e\|}$.
% The resulting orthonormal basis for the Krylov subspace is: $\{ \mathbf{q}_1, \mathbf{q}_2, \dots, \mathbf{q}_a \}
% $. The Krylov matrix is defined by
% \begin{equation}
%     V^n_a=\begin{bmatrix}
%         \mathbf{q}_1 & \mathbf{q}_2 & \cdots & \mathbf{q}_a
%     \end{bmatrix}.
% \end{equation}
This design choice is crucial because different timesteps $t_n$ have different system matrices $A^{n}$.

Based on this timestep-specific subspace design, our method operates in two distinct phases to achieve computational acceleration.
\begin{figure}[h!]
    \centering
    \includegraphics[width=1\linewidth]{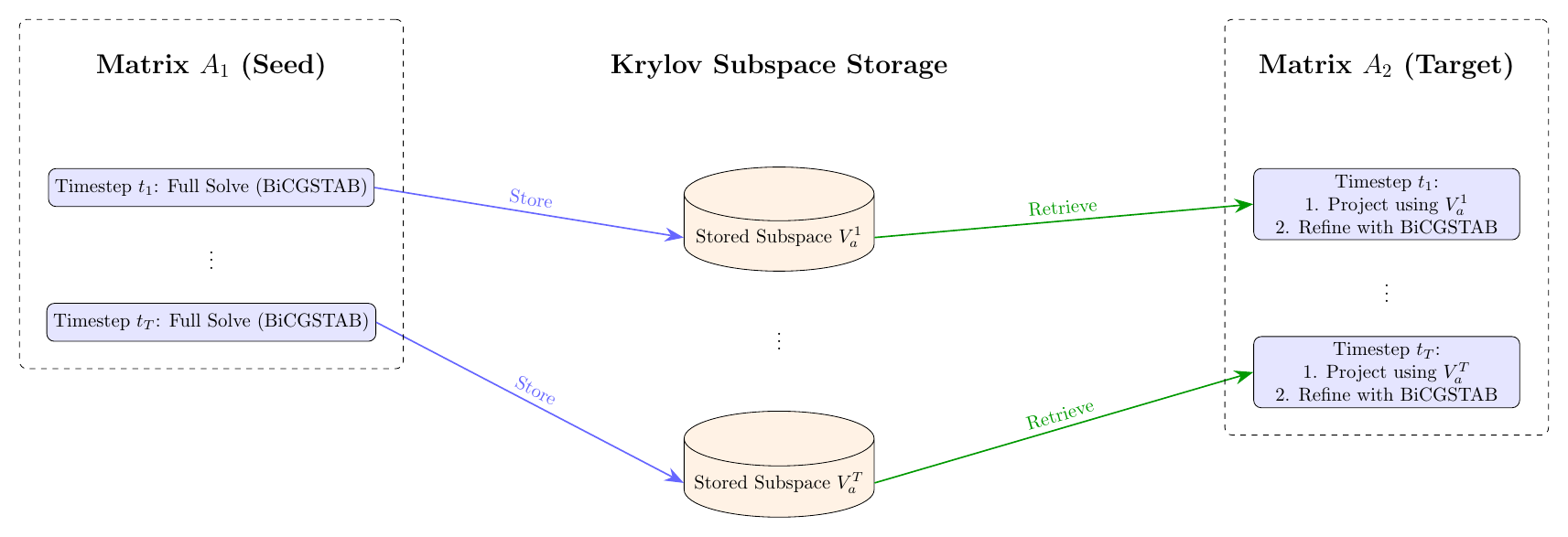}
    \caption{Conceptual diagram of our time-wise projection method. We consider all timesteps in the matrix $A^n_i$ as a seed; for each timestep, we use the projection method.}
\end{figure}
\subsubsection*{Phase 1: Seed Matrix Processing and Subspace Construction}

We define $A^n_1$ as the first seed system matrix $A^n_{\text{seed}}$. The first phase establishes the computational foundation by processing the seed matrix. We solve $A_{\text{seed}}^{n}m^{n} = b^{n}_{\text{seed}}$ using biconjugate gradient stabilized method (BiCGSTAB) \cite{van1992bi} to the desired tolerance $\epsilon$. BiCGSTAB begins with an initial guess and progressively refines it. This refinement is driven by several key vectors that are updated at each iteration. Central to this process is the residual vector $r^{n,(k)}$, which measures the error of the current guess $(r^{n,(k)}=b^{n,(k)}-A^{n,(k)}m^{n,(k)})$. The algorithm uses this residual to compute a search direction vector $p^{n,(k)}$, which points towards a better solution. Finally, a solution increment vector is calculated to update the current guess, and the process repeats until the error is acceptably small, less than tolerance $\epsilon$.

\subsubsection*{Phase 2: Target Matrix Acceleration via Projection}

The second phase leverages this stored cross-matrix Krylov subspace $V^n_a$ to accelerate subsequent system equations. For each target matrix $A^n_i$ where $i \in \{2, 3, \ldots, N\}$ and each timestep $t_n$, we perform timestep-wise projection using the corresponding matrix: $A_{\text{proj}}=(V^n_a)^TA_{\text{seed}}^{n} V^n_a \in \mathbb{R}^{a \times a}$ and $b_{\text{proj}} = (V^n_a)^T b_i^{n} \in \mathbb{R}^n$, where $a \ll HW$.
%(typically $a \approx 20$ and $HW \approx 16384$ for $128 \times 128 $ images).
Here we use $A_{\text{seed}}^n$ instead of the original system matrix $A^n_i$. This simplification makes sense since we assume different matrices are similar, as the corresponding images are similar, so that we can avoid the high computational cost of computing the projected matrix $(V^n_a)^TA_i^{n} V^n_a$ for each $i$. This savings is significant when $N$ is large. Also, the reduction in problem size enables efficient solution of the reduced system $A_{\text{proj}} \alpha = b_{\text{proj}}$. The solution is then reconstructed as an initial guess $x_0 = V^n_k \alpha$ \cite{chan1999galerkin} for the iterative method for solving the full system $A^{n}_i m^{n} = b_i^{n}$. 
%Furthermore, we enhance the Krylov subspace by periodically incorporating information from a new seed matrix. We accumulate the key vectors from each new seed for every $h$ system matrices. This creates an increasingly rich basis for accelerating future calculations.

\section{Image-by-Image Training for Generative Model with Score Embedding Framework}

Our primary goal is to increase the training efficiency of score-based diffusion models, which are typically slow to train for the traditional diffusion model, as they involve SDE in both the forward and reverse processes. To avoid this difficulty, we introduce a pre-training step where the score function $\nabla_x m$ is calculated numerically before the main training begins. This is done by solving the log-density Fokker-Planck equation \eqref{eq:log_fp} with a finite difference method in section \ref{finite difference method}. Then, we embed directly into the image data $x$ by solving the deterministic probability flow ODE:
\begin{equation}
    \frac{dx}{dt} = f(x,t) - \frac{1}{2}g^2(t)\nabla_x m.
    \label{probability flow ODE}
\end{equation}

It modifies the images so that they contain information about the diffusion dynamics. As a result, when the neural network is trained on these score-embedded images, its task is simplified. It does not have to learn the complex data distribution from scratch like the traditional diffusion model.

\subsection{Transport Equation for Score Embedding}

To embed it into the image data using the probability flow ODE \eqref{probability flow ODE} after computing the score function $\nabla_x m$, we do discretization on Eq.\eqref{probability flow ODE} with the same timestep total $T$, as we did discretization for Eq.\eqref{eq:log_fp} in section $\ref{finite difference method}$. The discretization would be $ x^{n} = x^{n-1}+\alpha^{n-1} \Delta t$, where $x^n$ refers to image data at the timestep $t_n$ and $ \alpha^{n-1}=f^{n-1}-\frac{1}{2}(g^{n-1})^2 \nabla_x m^{n-1}$.
This deterministic process generates a sequence of score-embedded images that capture the underlying diffusion dynamics.

\subsection{Training with Score Embedding}

The embedded images serve as enhanced training data. We use a modified U-Net architecture \cite{ronneberger2015u} with attention mechanisms \cite{vaswani2017attention}, training using the Adam optimizer \cite{kinga2015method} with the denoising objective \cite{boffi2023probability,zhang2023dive}:
\begin{equation}
\mathcal{L} = \mathbb{E}_{t,x,z} \left[ \frac{\|s_\theta(x + \lambda(t)z, t) + z/\lambda(t)\|^2}{2g(t)^2} \right],
\label{loss function}
\end{equation}
where $\lambda(t)$ is the perturbation parameter, $s_\theta(x + \lambda(t)z, t)$ represents the neural network, and $z$ is random noise. If we assume the loss function \eqref{loss function} goes to $0$, we can multiply $\lambda(t)$ into the loss function. This makes the training efficient and stable, since the term $z/\lambda(t)$ becomes very large and causes unstable gradients as $\lambda(t)$ is usually a small number \cite{ho2020denoising}. The final loss function becomes:
\begin{equation}
    \mathcal{L} = \mathbb{E}_{t,x,z} \left[ \frac{\|s_\theta(x + \lambda(t)z, t)\lambda(t) + z\|^2}{2g(t)^2} \right].
\end{equation}

Our training follows an image-by-image approach with SSIM-based early stopping. For each image, we train until reaching a target satisfactory structural similarity index measure (SSIM) threshold, then proceed to the next image. This strategy ensures high-quality learning while avoiding overfitting and over-training.

New images are generated through reverse SDE sampling \cite{song2020score,anderson1982reverse}, starting from pure noise and iteratively denoising using the trained score function. The process follows the equation: 
\begin{equation}
     x^{(n-1)} = x^{(n)} +s_\theta^* \cdot g(t)^2 \cdot \Delta t + g(t) \sqrt{ \Delta t} \cdot z ,
\end{equation}
where the $s_\theta^*$ is the score prediction that comes from our trained network, and the process continues for a specified number of denoising steps. We train the model until it achieves a SSIM \cite{wang2004image}, so that the training process is efficient. This training framework successfully reduces training time compared to traditional score-based models, as the training task is simplified.

\section{Experiment}

We conduct comprehensive experiments to validate the pre-computed score training and our cross-image Krylov projection method across three key dimensions: (1) computational efficiency gains in score pre-computation compared to DDPM, (2) acceleration benefits compared to standard sparse solvers, and (3) generation quality under identical computational budgets compared to DDPM. All experiments were run on a system with a MacBook Air M4 CPU and an Nvidia V100 GPU. The score pre-computation was performed on the CPU, while the neural network training was carried out on the GPU. We evaluate our approach on standard benchmark datasets: CIFAR-10 \cite{krizhevsky2009learning} and CelebA \cite{liu2015deep}. We use CIFAR-10 at a resolution of 32×32 pixels and CelebA at 64×64 and 128×128 pixels. Our approach uses the following configuration in all of our experiments:  timesteps $T=100$, diffusion coefficient $g(t) = 0.5$, drift term $f(t) = \vec{0}$, perturbation parameter $\lambda(t) = 0.1$, and a learning rate = $1 \times 10^{-3} $. We begin with a direct comparison against established baselines.

\subsection{Experiment 1: Efficiency Comparison with DDPM on Single Image Denoising}

In this experiment, we demonstrate the efficiency gains of our proposed method. We train each model by adding equivalent random noise to the same single image and train until they reach a specified Structural Similarity Index Measure (SSIM), then report their Mean Squared Error (MSE), Peak Signal-to-Noise Ratio (PSNR), \cite{gray1976review}, and training time. The training time for our method is the sum of both the score computation time, which looks for $\nabla_x m$, before the training process and the standard neural network training time to learn. 
% The mean square error measures the average squared difference between the pixel values of the original image $O$ and the reconstructed image $R$. A lower MSE indicates higher fidelity. For images of size $H \times W$, MSE is defined as:
% \begin{equation}
%     \text{MSE}=\dfrac{1}{HW}\sum_{i=1}^{H}\sum_{j=1}^{W}[O(i,j)-R(i,j)]^2 ,
% \end{equation}
% where $O(i,j)$ is the pixel value at row $i$ and column $j$ in the image $O$ and $R(i,j)$ is the pixel value at row $i$ and column $j$ in the image $R$. We also want to measures the ratio between the maximum possible power of a signal (image) and the power of corrupting noise that affects its quality by Peak Signal-to-Noise Ratio (PSNR). It is expressed in decibels (dB), and a higher PSNR generally indicates better reconstruction quality \cite{chanda2011digital}. It is defined using the MSE as:
% \begin{equation}
%     \text{PSNR}=10\times\log_{10}\left( \frac{\text{MAX}^2}{\text{MSE}} \right),
% \end{equation}
% where MAX is the maximum possible pixel value of the image. Speedup Factor quantifies the efficiency gain of our method relative to DDPM. It is calculated as the ratio of the DDPM's training time to our method's total training time.
 
\begin{table}[h!]
\centering
\begin{minipage}{0.48\textwidth}
\centering
\textbf{Target SSIM = 0.95}
\begin{tabular}{|l|c|c|}
\hline
\textbf{Metric} & \textbf{Our Method} & \textbf{DDPM} \\
\hline
Training time (s) &  16.75 &  356.68s \\
MSE &   0.002586 & 0.005398 \\
PSNR (dB) &  15.87  & 22.68 \\
Speedup Factor & \textbf{21.30×} & 1 \\
\hline
\end{tabular}
\end{minipage}
\hfill
\begin{minipage}{0.48\textwidth}
\centering
\textbf{Target SSIM = 0.99}
\begin{tabular}{|l|c|c|}
\hline
\textbf{Metric} & \textbf{Our Method} & \textbf{DDPM} \\
\hline
Training time (s) &  60.27 & 1634.93 \\
MSE &    0.000958 & 0.001122 \\
PSNR (dB) &  30.19 & 29.50 \\
Speedup Factor & \textbf{27.13×} & 1 \\
\hline
\end{tabular}
\end{minipage}
\caption{We tabulate the Training time (s) results for our method and DDPM on CIFAR-10 32×32.}

\centering
\begin{minipage}{0.48\textwidth}
\centering
\textbf{Target SSIM = 0.95}
\begin{tabular}{|l|c|c|}
\hline
\textbf{Metric} & \textbf{Our Method} & \textbf{DDPM} \\
\hline
Training time (s) &   80.04 & 5036.53 \\
MSE &    0.007052 & 0.008262 \\
PSNR (dB) & 21.52 & 20.83 \\
Speedup Factor & \textbf{62.93×} & 1 \\
\hline
\end{tabular}
\end{minipage}
\hfill
\begin{minipage}{0.48\textwidth}
\centering
\textbf{Target SSIM = 0.99}
\begin{tabular}{|l|c|c|}
\hline
\textbf{Metric} & \textbf{Our Method} & \textbf{DDPM} \\
\hline
Training time (s) &  122.45 & 14131.53\\
MSE &    0.001375 & 0.001340 \\
PSNR (dB) &  28.62 & 28.73 \\
Speedup Factor & \textbf{115.40×} & 1 \\
\hline
\end{tabular}
\end{minipage}
\caption{We tabulate the Training time (s) results for our method and DDPM on CelebA 64×64.}

\centering
\begin{minipage}{0.48\textwidth}
\centering
\textbf{Target SSIM = 0.85}
\begin{tabular}{|l|c|c|}
\hline
\textbf{Metric} & \textbf{Our Method} & \textbf{DDPM} \\
\hline
Training time (s) &   166.14 & 7329.93 \\
MSE &    0.014860 & 0.023668 \\
PSNR (dB) & 18.28 & 16.26 \\
Speedup Factor & \textbf{44.12×} & 1 \\
\hline
\end{tabular}
\end{minipage}
\hfill
\begin{minipage}{0.48\textwidth}
\centering
\textbf{Target SSIM = 0.90}
\begin{tabular}{|l|c|c|}
\hline
\textbf{Metric} & \textbf{Our Method} & \textbf{DDPM} \\
\hline
Training time (s) &  186.09s & 11319.86s\\
MSE &   0.009359 & 0.015459 \\
PSNR (dB) &  20.29 & 18.11 \\
Speedup Factor & \textbf{60.83×} & 1 \\
\hline
\end{tabular}
\end{minipage}
\caption{We tabulate the Training time (s) results for our method and DDPM on CelebA 128×128.}

\end{table}

% The experimental results, detailed in Tables 1-3, consistently demonstrate the superior efficiency and effectiveness of our proposed method compared to the DDPM baseline across all tested datasets and resolutions. Table 1 presents the comparison on the CIFAR-10 (32×32) dataset for two target SSIM values, 0.95 and 0.99. The rows report key metrics: total training time, final MSE, PSNR, and the speedup factor of our method over the baseline. At the SSIM target of 0.95, our method achieves the goal in just 16.75 seconds, a 21.30× speedup compared to the 356.68 seconds required by DDPM. This efficiency gain is even more pronounced at the higher quality target of 0.99, where our method achieves a 27.13× speedup. This trend of significant performance improvement continues on higher-resolution datasets. As shown in Table 2, on the CelebA (64×64) dataset, our method achieves massive speedups of 62.93× and 115.40× for SSIM targets of 0.95 and 0.99, respectively. In both scenarios, our method not only trains faster but also produces images of slightly better quality, evidenced by a lower MSE and a higher PSNR. Finally, the results on the CelebA (128×128) dataset in Table 3 further solidify these findings. For SSIM targets of 0.85 and 0.90, our method delivers substantial speedups of 44.12× and 60.83×. It is notable that even with these significant reductions in training time, our model produces images with demonstrably higher quality (lower MSE and higher PSNR) than the baseline.

The experimental results, detailed in Tables 1-3, consistently demonstrate the efficiency of our method over the DDPM baseline. For the CIFAR-10 (32×32) dataset, our method achieved a 21.30× speedup at an SSIM of 0.95 and a 27.13× speedup at an SSIM of 0.99, completing training in 16.75 and 60.27 seconds, respectively, compared to DDPM's 356.68 and 1634.93 seconds. The efficiency gains were even more significant on higher-resolution datasets, reaching speedups of 62.93× and 115.40× on CelebA (64×64) and 44.12× and 60.83× on CelebA (128×128). Across all resolutions, our method not only trained faster but also produced images with better quality, as indicated by a lower MSE and a higher PSNR when compared to the DDPM baseline.

This significant performance gain is attributed to the exact score supervision from our pre-computed Fokker-Planck solutions, which enables much faster convergence to high-quality results compared to DDPM’s noise-based training approach. The practical impact of this acceleration is substantial. For instance, on the CelebA 64×64 dataset, a training process that took the DDPM baseline over four hours was accomplished by our method in just over two minutes, without a discernible loss in final image quality.

The results demonstrate that our score embedding approach accelerates the learning process.

\begin{figure}[h!]
    \centering
    \includegraphics[width=0.9\linewidth]{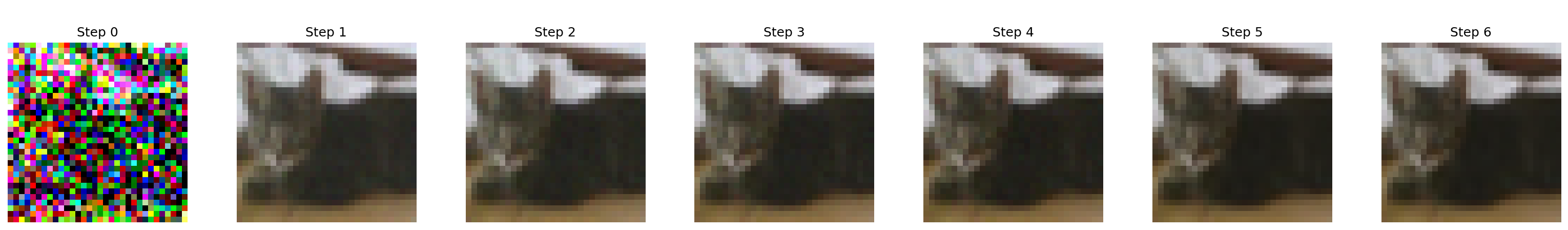}
    \includegraphics[width=0.9\linewidth]{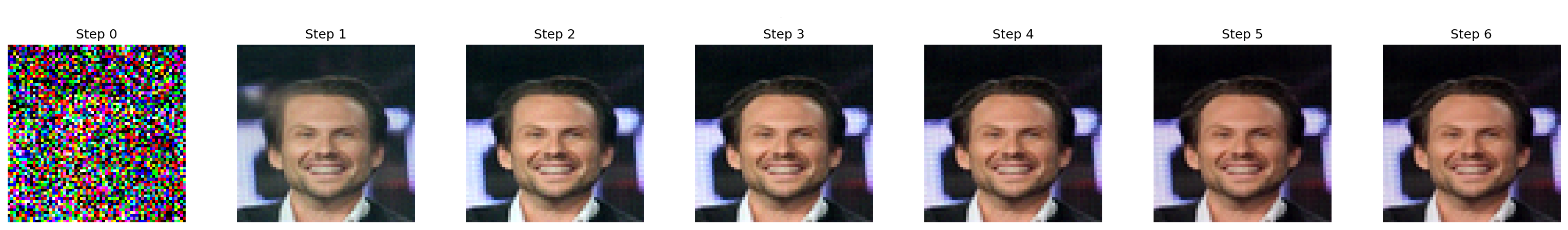}
    \includegraphics[width=0.9\linewidth]{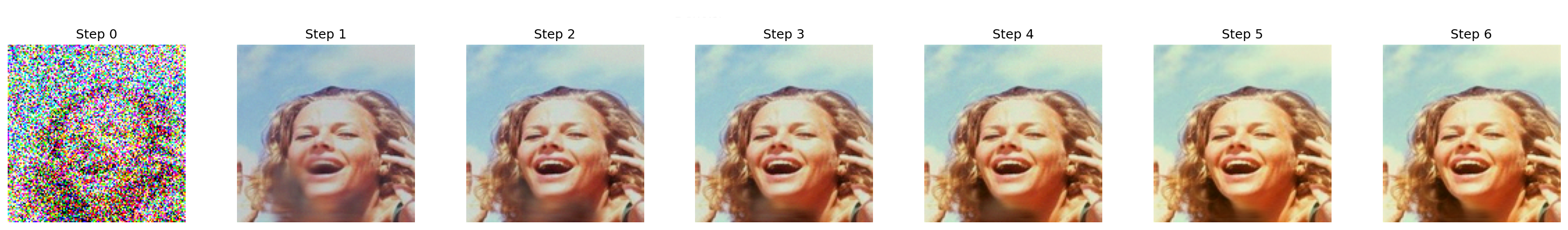}
    \caption{Demonstration of pre-computed Score Method for Denoising 32×32 (top), 64×64 (middle) and 128×128 (bottom).}
\end{figure}

\subsection{Experiment 2: Efficiency Comparison with Our Projection Method and Spsolve}
\label{Experiment 2}

In this section, we evaluate the computational efficiency of our cross-matrix Krylov projection method against direct solve for the sparse linear system solver for multiple systems of linear equations. The evaluation is performed on the task of pre-computation across 100 images from CIFAR-10 (Same category) and CelebA. This experiment isolates the performance gains achieved purely through our projection methodology, independent of neural network training considerations. The 50-image cycle strategy is employed for Krylov subspace expansion because it represents a trade-off between high memory and computational cost for other small image cycles, with subspace dimension limited to $a = 20$ vectors.

\begin{table}[h!]
\centering

\makebox[\linewidth][c]{\textbf{100 images on CIFAR-10 (32×32)}}
\begin{tabular}{@{}lcc@{}}
\toprule
\textbf{Metric} & \textbf{SpSolve} & \textbf{Our Projection Method} \\
\midrule
Total Time             & 51.684 s   & 43.500 s    \\
Average L2 Error       & 0          & 0.0558      \\
Total Time Reduction   & -          & \textbf{15.8\%} \\
\bottomrule
\end{tabular}

\par\medskip
\makebox[\linewidth][c]{\textbf{100 images on CelebA (64×64)}}
\begin{tabular}{@{}lcc@{}}
\toprule
\textbf{Metric} & \textbf{SpSolve} & \textbf{Our Projection Method} \\
\midrule
Total Time             & 244.742 s  & 171.927 s   \\
Average L2 Error       & 0          & 0.0767      \\
Total Time Reduction   & -          & \textbf{29.8\%} \\
\bottomrule
\end{tabular}

\par\medskip
\makebox[\linewidth][c]{\textbf{100 images on CelebA (128×128)}}
\begin{tabular}{@{}lcc@{}}
\toprule
\textbf{Metric} & \textbf{SpSolve} & \textbf{Our Projection Method} \\
\midrule
Total Time             & 1222.618 s & 688.749 s   \\
Average L2 Error       & 0          & 0.0950      \\
Total Time Reduction   & -          & \textbf{43.7\%} \\
\bottomrule
\end{tabular}

\par\medskip
\caption{Computational performance of Krylov projection method compared to the SpSolve baseline. The experiment was run on 100 images for three datasets of increasing resolution (CIFAR-10 32×32, CelebA 64×64, and CelebA 128×128). Time reduction is calculated as the percentage decrease in total time relative to the SpSolve baseline.}
\end{table}

To evaluate the computational efficiency of the Krylov projection method, we compare it against the standard SpSolve solver, a direct solver developed using python. Table 4 explicitly details this comparison, showing the performance of both methods on batches of 100 images across three distinct resolutions: CIFAR-10 (32×32), CelebA (64×64), and CelebA (128×128). For each dataset, we report the total time required, the average L2 relative error introduced by our projection method (for which SpSolve is the baseline), and the resulting overall time reduction.

Our method demonstrates consistent speedup across all datasets with increasing benefits for larger images. For CIFAR-10 (32×32), we achieve 15.8$\%$ time reduction (43.50s vs 51.68s total time). CelebA (64×64) shows 29.8$\%$ time reduction (171.92s vs 244.74s), while CelebA (128×128) achieves 43.7$\%$ time reduction (688.74s vs 1222.61s). Meanwhile, L2 norm relative errors remain consistently low across all datasets (0.0558–0.0950). This demonstrates that our projection-based acceleration maintains mathematical accuracy while providing computational benefits. The increasing time reduction (15.8$\%$ → 29.8$\%$ → 43.7$\%$) with image resolution shows that our cross-matrix Krylov approach becomes more effective for larger, computationally intensive problems. To illustrate the robustness of these findings, we repeated the experiment on other distinct sets of 100 images each, and the performance improvements remained consistent.

\subsection{Experiment 3: Image Quality Comparison Under a Fixed Computational Budget}
\label{Experiment 3}

We train for our image-by-image generative method and train for DDPM with the same training time. Then, we evaluate the quality achievable within the same computational constraints. This experiment addresses the question: given limited time resources, which method produces higher quality results?

% %32x32 generation
%     \begin{figure}[h!]
%         \centering
%         \includegraphics[width=0.8\linewidth]{progressive_generation_truck_1.png}

%         \includegraphics[width=0.8\linewidth]{progressive_generation_truck_2.png}

%         \includegraphics[width=0.8\linewidth]{progressive_generation_truck_3.png}
%         \caption{Demonstration of generation of 32×32 truck images from CIFAR-10. We sample 7 timesteps during the sampling to demonstrate the generating process.}
%     \end{figure}
%64x64 generation
    % \begin{figure}[h!]
    %     \centering
    %     \includegraphics[width=0.8\linewidth]{progessive_generation_celeba_64x64_1.png}

    %     \includegraphics[width=0.8\linewidth]{progessive_generation_celeba_64x64_2.png}

    %     \includegraphics[width=0.8\linewidth]{progessive_generation_celeba_64x64_3.png}
    %     \caption{Demonstration of generation of 64×64 celebrity images from CelebA. We sample 7 timesteps during the sampling to demonstrate the generating process.}
    % \end{figure}
%128x128 generation
    \begin{figure}[h!]
        \centering
        \includegraphics[width=0.9\linewidth]{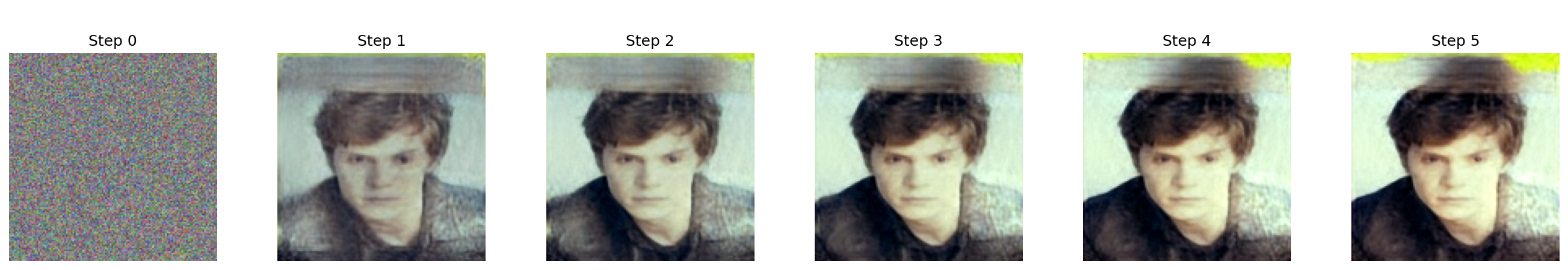}
        \includegraphics[width=0.9\linewidth]{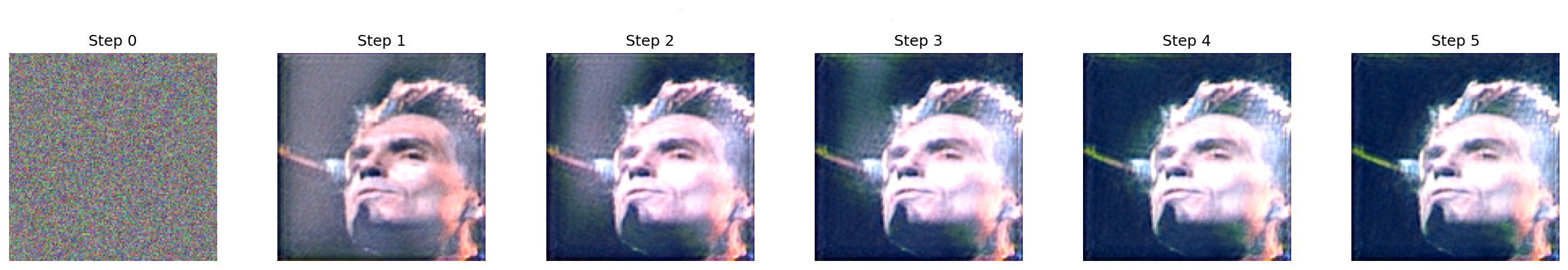}
        \includegraphics[width=0.9\linewidth]{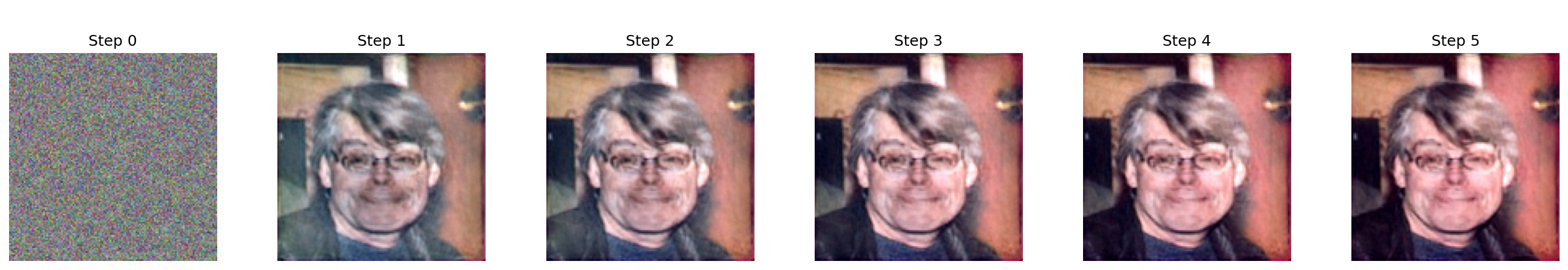}
        \caption{Demonstration of generation of 128×128 celebrity images from CelebA. We sample 6 timesteps during the sampling to demonstrate the generating process.}
    \end{figure}
The overall computational cost of our generative model is divided into two distinct phases. First, the Pre-compute Time represents the time cost of calculating the score function for a given image using the Krylov Projection method. Following this, the Training Time accounts for the iterative, image-by-image process of training the neural network. Table \ref{tab:computation_times} shows the pre-compute time, training time, and the resulting total time (the sum of the two) for generating images of different sizes: 32×32, 64×64, and 128×128.
\begin{table}[h!]
  \centering
  \begin{tabular}{lccc}
    \hline
    \textbf{Image Size} & \textbf{Pre-compute Time} & \textbf{Training Time} & \textbf{Total Time} \\
    \hline
    32x32 & 35.45 s & 4119.72 s & 4155.17 s \\
    64x64 & 128.56 s & 6260.03 s & 6388.59 s \\
    128x128 & 556.71 s & 28585.18 s & 29141.89 s \\
            \bottomrule
        \end{tabular}
    \caption{Pre-computation and Training Time of Our Generative Model}
    \label{tab:computation_times}
\end{table}

We compared the images generated by our method with those from DDPM when both were allocated comparable time budgets. Figures 4, 5, and 6 demonstrate that under a fixed time budget, our method (left panels) and the DDPM method (right panels) generate images of trucks and faces in different resolutions. 

\begin{figure}[h!]
    \centering
    \begin{subfigure}[b]{0.45\textwidth}
        \centering
        \includegraphics[width=\linewidth]{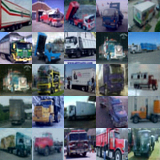}
        \label{fig:grid_b}
    \end{subfigure}
    \begin{subfigure}[b]{0.45\textwidth}
        \centering
        \includegraphics[width=\linewidth]{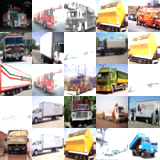}
        \label{fig:grid_a}
    \end{subfigure}
    \caption{A comparison of 32×32 images generated after an identical training time budget (4155.17s). Our method (left) produces coherent images with an average SSIM of 0.8991, whereas DDPM (right) achieves truck images with an average SSIM of only 0.8517 within the same time budget.}
\end{figure}

\begin{figure}[h!]
    \centering
    \begin{subfigure}[b]{0.45\textwidth}
        \centering
        \includegraphics[width=\linewidth]{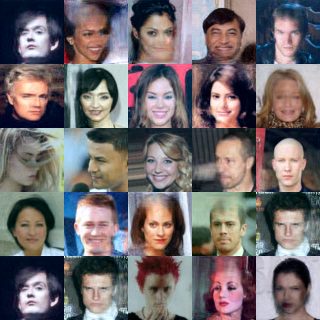}
        \label{fig:grid_b}
    \end{subfigure}
    \begin{subfigure}[b]{0.45\textwidth}
        \centering
        \includegraphics[width=\linewidth]{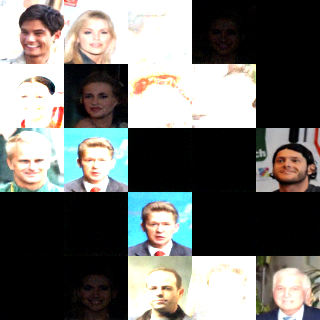}
        \label{fig:grid_a}
    \end{subfigure}
    \caption{A comparison of 64×64 images generated after an identical training time budget (6388.59s). Our method (left) produces coherent images with an average SSIM of 0.6302. In contrast, the standard DDPM (right) only produces a few recognizable images with an average SSIM of 0.4446 within the same time constraint.}
\end{figure}

\begin{figure}[h!]
    \centering
    \begin{subfigure}[b]{0.45\textwidth}
        \centering
        \includegraphics[width=\linewidth]{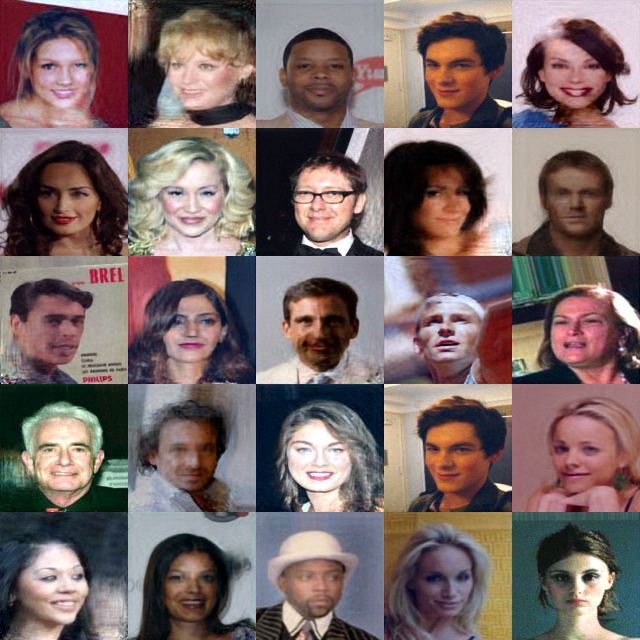}
        \label{fig:grid_b}
    \end{subfigure}
    \begin{subfigure}[b]{0.45\textwidth}
        \centering
        \includegraphics[width=\linewidth]{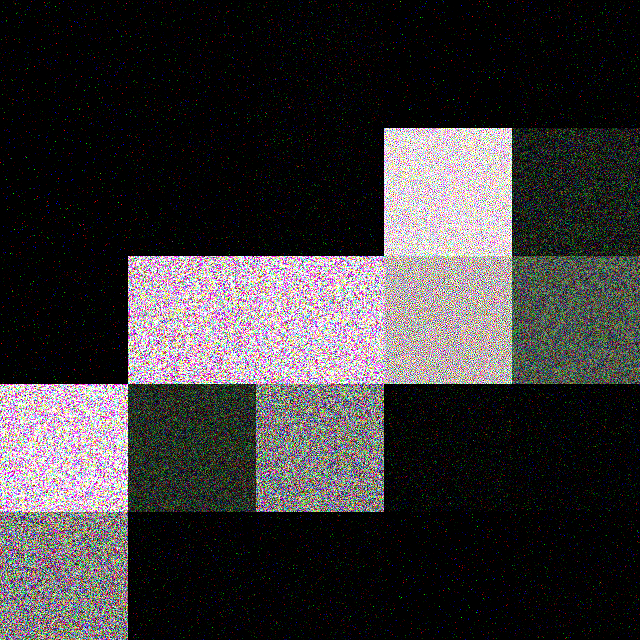}
        \label{fig:grid_a}
    \end{subfigure}
    \caption{A comparison of 128×128 images generated after an identical training time budget (29141.89s). Our method (left) produces coherent images with an average SSIM of 0.8312. In contrast, the standard DDPM (right) fails to produce any recognizable images within the same time constraint, highlighting our model's superior training efficiency.}
\end{figure}

We consistently achieve higher image quality metrics within comparable time budgets. The exact score supervision from pre-computed Fokker-Planck solutions enables faster convergence to high-quality results compared to DDPM's noise-based training approach. While DDPM requires extensive training time to reach target quality levels, our method reaches comparable or superior quality much earlier in the training process. The results demonstrate that our score embedding approach accelerates the learning process. 

Rather than learning score functions from scratch through denoising objectives, our method leverages mathematically precise scores to guide the network more efficiently. This leads to a higher final quality within fixed time constraints. For example, high-quality 32×32 images were generated in only 1.15 hours.

We chose to omit a comparative analysis using the Fréchet Inception Distance (FID) for several reasons. Primarily, under the strict computational budget of our experiments, the DDPM baseline failed to produce coherent images, generating largely noise. An FID score would therefore be uninformative, as it would simply measure the distance between our model’s coherent output and a distribution of noise. Furthermore, our training paradigm prioritizes rapid convergence to high per-sample fidelity—using SSIM and PSNR as quality guides—rather than explicitly modeling the global diversity of the entire dataset, making these metrics more aligned with our stated contributions. Finally, the practical requirement of generating tens of thousands of samples for a stable FID score was outside the scope of these resource-constrained experiments

% We omitted a comparative analysis using the Fréchet Inception Distance (FID) for several reasons. Primarily, the goal of Experiment 3 was to evaluate image quality under a strict computational budget. Under these constraints, the DDPM baseline fails to produce coherent images, generating outputs that are largely noise. A comparison of FID scores would therefore be uninformative, as it would merely measure the distance between our model’s coherent image distribution and a distribution of noise.

% Moreover, our training paradigm prioritizes rapid convergence to high per-sample fidelity—using SSIM as a quality guide—rather than explicitly modeling the global diversity of the entire dataset. Thus, metrics such as SSIM and PSNR are more aligned with our stated contributions. Finally, the practical requirement of generating tens of thousands of samples for a stable FID score was outside the scope of these resource-constrained experiments.

The proposed cross-image projection relies on structural similarities between matrices which are related to images; therefore, its effectiveness may be reduced when applied to datasets with extreme diversity. This difficulty could potentially be resolved by using clustering to create distinct seed images for different image groups.

Ultimately, this experiment validates the real-world utility of our approach for time-constrained scenarios. This makes our method a practical solution for organizations with limited computational budgets, allowing them to achieve better results and making high-quality image generation more accessible and cost-effective.

\section{Conclusion}

This paper introduces a new framework to speed up score-based diffusion models using two main innovations: Cross-matrix Krylov projection for efficient score pre-computation and an image-by-image training approach. The cross-matrix Krylov projection method exploits the structural similarities between system matrices across different images. By using a "seed" matrix to build a shared Krylov subspace, it can quickly solve for subsequent "target" matrices. This technique reduces pre-computation time by 15.8\% to 43.7\% compared to standard sparse solvers, with minimal loss of accuracy. The framework also uses a pre-computed score embedding, which simplifies the training task for the neural network. This, combined with an adaptive, SSIM-based early-stopping strategy, significantly accelerates the training process. In experiments, this method achieved a total speedup of 21.30× to 115.40× over DDPM baselines for single-image denoising. The training time was reduced from thousands of seconds to under 400 seconds, with even greater benefits for higher-resolution images. Furthermore, under a fixed computational budget, this framework consistently produced coherent, high-quality images, while DDPM often failed to generate recognizable content. In short, this work bridges numerical methods with generative modeling to make high-quality image generation more accessible and cost-effective for users with limited computational resources.

\bibliographystyle{unsrt}
\bibliography{references}

\section{Appendix}

\subsection{Full Derivation of the Fokker-Planck equation and discretization to form the system of linear equations}

\label{Full Derivation of the Fokker-Planck equation and discretization to form the system of linear equations}
Stochastic Differential Equation is given by:
\begin{equation}
     dx(t) = f(x,t)dt + g(t)W(t),
\end{equation}
where
\begin{align}
 x(t): \mathbb{R} &\to \mathbb{R}^D ,\\
 f(x,t): \mathbb{R}^{D+1} &\to \mathbb{R}^D ,\\
 g(t): \mathbb{R} &\to \mathbb{R}^{D \times D}, \\
 W(t): \mathbb{R} &\to \mathbb{R}^D ,
\end{align}
for some $D \in \mathbb{N}$, note $g(t)$ can be written as a scalar.

\paragraph{Fokker-Planck Equations a.k.a. Kolmogorov Forward Equation}
The equation is given by Bernt Øksendal \cite{oksendal2013stochastic} on page 182.
\begin{equation}
    \begin{split}
        \frac{\partial}{\partial t} p(x,t) = - \sum_{i=1}^{D} \frac{\partial}{\partial x_i} [f_i(x,t) p(x,t)] 
        + \sum_{i=1}^{D} \sum_{j=1}^{D} \frac{\partial^2}{\partial x_i \partial x_j} [\textbf{D}_{ij}(x,t) p(x,t)],
    \end{split}
\end{equation}
where the Tensor $\textbf{D} := \frac{1}{2} g(t) g(t)^T = \frac{1}{2} g^2(t)$ and $f(x,t) = (f_1, f_2, f_3, \dots, f_N) $.

\begin{align*}
\frac{\partial}{\partial t} p(x,t) &= - \sum_{i=1}^{N} \frac{\partial}{\partial x_i} [f_i(x,t) p(x,t)] 
 + \sum_{i=1}^{N} \sum_{j=1}^{N} \frac{\partial^2}{\partial x_i \partial x_j} \left[ \frac{1}{2} g^2(t) p(x,t) \right] \\
&= - \sum_{i=1}^{N} \frac{\partial}{\partial x_i} [f_i(x,t) p(x,t)] 
 + \frac{1}{2} g^2(t) \sum_{i=1}^{N} \sum_{j=1}^{N} \frac{\partial^2}{\partial x_i \partial x_j} p(x,t) .
\end{align*}
Note $p(x,t)$ can be written as $\prod_{k=1}^{D} p_k(x_k(t))$, as $\{ x_i \}_1^D$ is independent and identically distributed random variables.
\begin{align*}
  \sum_{i=1}^{D} \sum_{j=1}^{D} \frac{\partial^2}{\partial x_i \partial x_j} p(x,t) 
&= \sum_{i=1}^{D} \sum_{j=1}^{D} \frac{\partial^2}{\partial x_i \partial x_j} \prod_{k=1}^{D} p_k(x_k(t)) \\
&= \sum_{i=1}^{D} \sum_{j=1}^{D} \frac{\partial^2}{\partial x_i \partial x_j} p_1(x_1(t)) \cdots p_i(x_i(t)) \cdots p_D(x_D(t)) \\
&= \sum_{i=1}^{D} p_1(x_1(t)) \cdots \left( \frac{\partial^2}{\partial x_i^2} p_i(x_i(t)) \right) \cdots p_D(x_D(t)) \\
&= \sum_{i=1}^{D} \left( p_1(x_1(t)) \cdots \left( \frac{\partial^2}{\partial x_i^2} p_i(x_i(t)) \right) \cdots p_D(x_D(t)) \right) \cdot \frac{p_i(x_i(t))}{p_i(x_i(t))} \\
&= \sum_{i=1}^{D} p(x,t) \left( \frac{\partial^2}{\partial x_i^2} p_i(x_i(t)) \right) \cdot \frac{1}{p_i(x_i(t))} \\
&= p(x,t) \sum_{i=1}^{D} \left( \frac{\partial^2}{\partial x_i^2} p_i(x_i(t)) \right) \cdot \frac{1}{p_i(x_i(t))} .
\end{align*}
Note that $ \frac{\partial}{\partial x_i} \log p_i(x_i(t)) = \frac{1}{p_i(x_i(t))} \frac{\partial}{\partial x_i} p_i(x_i(t)) $.
\\ $\implies \frac{\partial^2}{\partial x_i^2} \log p_i(x_i(t)) = - \frac{1}{(p_i(x_i(t)))^2} \left( \frac{\partial}{\partial x_i} p_i(x_i(t)) \right)^2 + \frac{1}{p_i(x_i(t))} \frac{\partial^2}{\partial x_i^2} p_i(x_i(t)) $.
\\ Denote $\nabla^2$ as Laplacian Operator and hence,
\begin{align*}
    \sum_{i=1}^{D} \sum_{j=1}^{D} \frac{\partial^2}{\partial x_i \partial x_j} p(x,t)  & = p(x,t) \sum_{i=1}^{D} \left( \frac{\partial^2}{\partial x_i^2} \log p_i(x_i(t)) \right)  + p(x,t) \sum_{i=1}^{D}\left( \frac{1}{(p_i(x_i(t)))^2} \left( \frac{\partial}{\partial x_i} p_i(x_i(t)) \right)^2 \right) \\
&= p(x,t) \nabla^2 \log p(x,t) + p(x,t) \sum_{i=1}^{D} \left( \frac{1}{p_i(x_i(t))} \frac{\partial}{\partial x_i} p_i(x_i(t)) \right)^2 \\
&= p(x,t) \nabla^2 \log p(x,t) + p(x,t) \sum_{i=1}^{D} \left( \frac{\partial}{\partial x_i} \log p_i(x_i(t)) \right)^2 .
\end{align*}
Then,
\begin{align*}
\frac{\partial}{\partial t} p(x, t) = p_t(x, t) 
=&- \sum_{i=1}^{D} \frac{\partial}{\partial x_i} [f_i(x, t) p(x, t)] + \frac{1}{2} g^2(t) [p(x, t) \nabla^2 \log p(x, t)] + \frac{1}{2} g^2(t) p(x, t) \langle \nabla_x \log p(x, t), \nabla_x \log p(x, t) \rangle \\
 = &- \sum_{i=1}^{D} \frac{\partial}{\partial x_i} [f_i(x, t) p(x, t)] + \frac{1}{2} g^2(t) [p(x, t) \nabla^2 \log p(x, t)] + \frac{1}{2} g^2(t) p(x, t) \langle \nabla_x \log p(x, t), \nabla_x \log p(x, t) \rangle .
\end{align*}
We use the product rule of partial derivatives,
\begin{align*}
p_t(x, t) = &- \sum_{i=1}^{D} p(x, t) \frac{\partial}{\partial x_i} f_i(x, t) - \sum_{i=1}^{D} f_i(x, t) \frac{\partial}{\partial x_i} p(x, t)  \frac{1}{2} g^2(t) [p(x, t) \nabla^2 \log p(x, t)] \frac{1}{2} g^2(t) p(x, t) \langle \nabla_x \log p(x, t), \nabla_x \log p(x, t) \rangle \\
= &-p(x, t) \cdot \text{Div} f(x, t) -\sum_{i=1}^{D} f_i(x, t) p_1(x_1(t)) \cdots \frac{\partial}{\partial x_i} p_i(x_i(t)) \cdots p_D(x_D) \\ & + \frac{1}{2} g^2(t) [p(x, t) \nabla^2 \log p(x, t)] + \frac{1}{2} g^2(t) p(x, t) \langle \nabla_x \log p(x, t), \nabla_x \log p(x, t) \rangle .
\end{align*}
Note $\frac{\partial}{\partial x_i} \log p_i(x_i(t)) = \frac{1}{p_i(x_i(t))} \frac{\partial}{\partial x_i} p_i(x_i(t)), \forall i \in \{1, \dots, D\}$. Then, 
\begin{align*}
& \sum_{i=1}^{D} f_i(x, t) p_1(x_1(t)) \cdots \frac{\partial}{\partial x_i} p_i(x_i(t)) \cdots p_D(x_D(t)) \\
= &\sum_{i=1}^{D} f_i(x, t) p_1(x_1(t)) \cdots p_i(x_i(t)) \frac{\partial}{\partial x_i} \log p_i(x_i(t)) \cdots p_D(x_D(t)) \\
=& \sum_{i=1}^{D} f_i(x, t) \frac{\partial}{\partial x_i} \log p_i(x_i(t)) p(x, t) \\
=& p(x, t) \sum_{i=1}^{D} f_i(x, t) \frac{\partial}{\partial x_i} \log p_i(x_i(t))
\end{align*}.

Also note that
\begin{align*}
& \sum_{i=1}^{D} f_i(x, t) \frac{\partial}{\partial x_i} \log p_i(x_i(t)) \\ &=  (f_1(x, t),f_2(x, t),f_3(x, t), \cdots, f_D(x, t))  \cdot (\frac{\partial}{\partial x_1}\log{p_1(x_1,t)},\frac{\partial}{\partial x_2}\log{p_2(x_2,t)},\cdots,\frac{\partial}{\partial x_i}\log{p_D(x_D,t)}) .
\end{align*}
Then,
\begin{align*}
 & p_t(x, t) \\
=& -p(x, t) \cdot \text{Div} f(x, t)  - \sum_{i=1}^{D} f_i(x, t) p_1(x_1(t)) \cdots \frac{\partial}{\partial x_i} p_i(x_i(t)) \cdots p_D(x_D)  + \frac{1}{2} g^2(t) [p(x, t) \nabla^2 \log p(x, t)] \\
& + \frac{1}{2} g^2(t) p(x, t) \langle \nabla_x \log p(x, t), \nabla_x \log p(x, t) \rangle \\
=& -p(x, t) \cdot \text{Div} f(x, t) - p(x, t) \sum_{i=1}^{D} f_i(x, t) \frac{\partial}{\partial x_i} \log p_i(x_i(t))  \\ &+ \frac{1}{2} g^2(t) [p(x, t) \nabla^2 \log p(x, t)]  + \frac{1}{2} g^2(t) p(x, t) \langle \nabla_x \log p(x, t), \nabla_x \log p(x, t) \rangle \\
=& -p(x, t) \cdot \text{Div} f(x, t) - p(x, t) \left( f(x, t) \cdot \nabla_x \log p(x, t) \right) + \frac{1}{2} g^2(t) [p(x, t) \nabla^2 \log p(x, t)] \\ & + \frac{1}{2} g^2(t) p(x, t) \langle \nabla_x \log p(x, t), \nabla_x \log p(x, t) \rangle .
\end{align*}
Hence,
\begin{align*}
p_t(x, t) = &-p(x, t) \cdot \text{Div} f(x, t) - \left(f(x, t) \cdot \nabla_x \log p(x, t) \right) p(x, t) \\
&+ \frac{1}{2} g^2(t) [p(x, t) \nabla^2 \log p(x, t)] + \frac{1}{2} g^2(t) p(x, t) \langle \nabla_x \log p(x, t), \nabla_x \log p(x, t) \rangle .
\end{align*}
For simplicity, we ignore variables $(x,t)$. It becomes
\begin{equation}
    \begin{split}
        p_t = -p \cdot \text{Div} f - p \langle f, \nabla_x \log p \rangle + \frac{1}{2} g^2 p \nabla^2 \log p + \frac{1}{2} g^2 p \langle \nabla_x \log p, \nabla_x \log p \rangle .
    \end{split}
\end{equation}
The Probability Flow ODE is as follow:
\begin{equation}
    dx = \left[ f(x, t) - \frac{1}{2} g^2(t) \nabla_x \log p(x, t) \right]dt .
\end{equation}
Let $m := \log p$, then $m_t = \frac{1}{p} p_t$,
\begin{equation}
    m_t = - \text{Div} f - \langle f, \nabla_x m \rangle + \frac{1}{2} g^2 \nabla^2 m + \frac{1}{2} g^2 \langle \nabla_x m, \nabla_x m \rangle .
    \label{FP}
\end{equation}
We do linear approximation of $\nabla_x m$, $\nabla_x \tilde{ m} \approx \nabla_x m$. The Eq.\eqref{FP} becomes:
\begin{equation}
    \frac{\partial m}{\partial t} = - \text{Div} f + \frac{1}{2} g^2 \nabla^2 m - \langle f, \nabla_x m \rangle + \frac{1}{2} g^2 \langle (\nabla_x \tilde{m}), \nabla_x m \rangle .
\end{equation}

\paragraph{Standard Five-point Stencil}
Let $t_i$ be the timestep, for all $i \in \mathbb{N}$, such that $t_i=(i-1)(\Delta t)(t) $, where $\Delta t = \frac{1}{T}$ is the temporal step size, and $T$ is the total timesteps. Let $x_i:=(i-1)(\Delta t)(x)$ and $y_i:=(i-1)(\Delta t)(y)$ be the discretized images over the pixels of each channel. Note that we use zero padding at the borders of the image as the boundary condition. Let $\Delta x = \Delta y = \frac{1}{H} = \frac{1}{W} =h$ be the spatial grid spacing. We discretize the log-density distribution as follows:

\begin{equation}
    m_{i,j}^n := m(x_i,y_j,t_n),
\end{equation}

\begin{equation}
    f^n_{i,j} := f(x_i,y_j,t_n),
\end{equation}
        
\begin{equation}
    g^n := g(t_n),
\end{equation}

\begin{equation}
        m_t \approx \frac{m_{i,j}^n - m_{i,j}^{n-1}}{\Delta t},
        \label{mt}
\end{equation}

\begin{equation}
    \nabla_x m \approx \left( \frac{m_{i+1,j}^n - m_{i-1,j}^n}{2h}, \frac{+ m_{i,j+1}^n - m_{i,j-1}^n}{2\Delta y} \right),
    \label{gram}
\end{equation}

\begin{equation}
    \nabla^2 m \approx \frac{m_{i+1,j}^n + m_{i,j+1}^n - 4m_{i,j}^n + m_{i-1,j}^n + m_{i,j-1}^n}{(h)^2} ,
    \label{gragram}
\end{equation}

\begin{equation}
        \text{Div} f \approx \frac{(f_{i+1,j} - f_{i-1,j})_x}{2 h} + \frac{(f_{i,j+1} - f_{i,j-1})_y}{2\Delta y}
        \label{divf},
\end{equation}where 
$f^n_{i,j}=((f^n_{i,j})_x,(f^n_{i,j})_y)$.
Recall $\overset{\eqref{mt}}{\overbrace{  \frac{\partial m}{\partial t }}} =
    - \overset{\eqref{divf}} {\overbrace{ \text{Div}f}} +
    \frac{1}{2} g^2 \cdot    \overset{\eqref{gragram}}{\overbrace{ \nabla^2 m}}
    - f \cdot     \overset{\eqref{gram}}{\overbrace{  \nabla_x m} }+
    \frac{1}{2} g^2  (\nabla_x \tilde{m})  \cdot     \overset{\eqref{gram}}{\overbrace{ (\nabla_x m)}} $.
We have   
\begin{align*}
     & \frac{m_{i,j}^n - m_{i,j}^{n-1}}{\Delta t} \\ = & -\frac{(f^n_{i+1,j})_x - (f^n_{i-1,j})_x}{2 h} - \frac{(f^n_{i,j+1})_y - (f^n_{i,j-1})_y}{2\Delta y}  \frac{1}{2} (g^n)^2 \left( \frac{m_{i+1,j}^n + m_{i,j+1}^n - 4m_{i,j}^n + m_{i-1,j}^n + m_{i,j-1}^n}{h^2} \right) \\
         &-  f^n \cdot \left( \frac{m_{i+1,j}^n - m_{i-1,j}^n}{2 h}, \frac{m_{i,j+1}^n - m_{i,j-1}^n}{2 \Delta y} \right) + \frac{1}{2} (g^n)^2 \nabla_x \tilde{m} \cdot \left( \frac{m_{i+1,j}^n - m_{i-1,j}^n}{2 h}, \frac{m_{i,j+1}^n - m_{i,j-1}^n}{2 h} \right) .
\end{align*}
Then, 
\begin{align*}
        & \frac{m_{i,j}^n}{\Delta t} - \frac{1}{2} (g^n)^2 \frac{m_{i+1,j}^n + m_{i,j+1}^n - 4m_{i,j}^n + m_{i-1,j}^n + m_{i,j-1}^n}{(h)^2}    + \left( f^n - \frac{1}{2} (g^n)^2 \nabla_x \tilde{m} \right) \cdot \left( \frac{m_{i+1,j}^n - m_{i-1,j}^n}{2 h}, \frac{ m_{i,j+1}^n - m_{i,j-1}^n}{2 h} \right) \\
        &= \frac{m_{i,j}^{n-1}}{\Delta t} - \frac{(f^n_{i+1,j} - f^n_{i-1,j})_x + (f^n_{i,j+1} - f^n_{i,j-1})_y}{2 h} .
    \end{align*}
Then,
\begin{align*}
        &   \frac{m_{i,j}^{n-1}}{\Delta t} - \frac{(f^n_{i+1,j} - f^n_{i-1,j})_x + (f^n_{i,j+1} - f^n_{i,j-1})_y}{2h} = \frac{m_{i,j}^n}{\Delta t} - \frac{1}{2} (g^n)^2 \left( \frac{ m_{i+1,j}^n + m_{i,j+1}^n - 4m_{i,j}^n + m_{i-1,j}^n + m_{i,j-1}^n }{h^2} \right) \\
        &  + \left( \left((f^n_{i,j})_x,(f^n_{i,j})_y \right) - \frac{1}{2} (g^n)^2 \left( \frac{\tilde{m}_{i+1,j}^n - \tilde{m}_{i-1,j}^n}{2h}, \frac{\tilde{m}_{i,j+1}^n - \tilde{m}_{i,j-1}^n}{2h} \right) \right) \cdot \left( \frac{m_{i+1,j}^n - m_{i-1,j}^n}{2h}, \frac{m_{i,j+1}^n - m_{i,j-1}^n}{2h} \right) .
\end{align*}
Eventually,
\begin{align*}
        & + \left(\frac{1}{\Delta t} + \frac{2(g^n)^2}{(h)^2}\right)m^n_{i,j} \\
        & + \frac{1}{2h}\left(-\frac{(g^n)^2}{h} + (f^n_{i,j})_x - \frac{1}{4 h}(g^n)^2 (\tilde{m}_{i+1,j}^n - \tilde{m}_{i-1,j}^n) \right) m^n_{i+1,j} \\
        & + \frac{1}{2h}\left(-\frac{(g^n)^2}{h}  + (f^n_{i,j})_y - \frac{1}{4 h}(g^n)^2 (\tilde{m}_{i,j+1}^n - \tilde{m}_{i,j-1}^n) \right) m^n_{i,j+1} \\
        & + \frac{1}{2h}\left(-\frac{(g^n)^2}{h}  - (f^n_{i,j})_x + \frac{1}{4 h}(g^n)^2 (\tilde{m}_{i+1,j}^n - \tilde{m}_{i-1,j}^n) \right) m^n_{i-1,j} \\
        & + \frac{1}{2h}\left(-\frac{(g^n)^2}{h}  - (f^n_{i,j})_y + \frac{1}{4 h}(g^n)^2 (\tilde{m}_{i,j+1}^n - \tilde{m}_{i,j-1}^n) \right) m^n_{i,j-1} \\
        & = \frac{m^{n-1}_{i,j}}{\Delta t} - \frac{(f_{i+1,j} - f_{i-1,j})_x + (f_{i,j+1} - f_{i,j-1})_y}{2 h},
    \end{align*}
where $f_{i,j}=((f^n_{i,j})_x,(f^n_{i,j})_y)$.

\paragraph{Matrix Representation of the Finite Difference Equation}
The system of equations, one for each grid point $(i,j)$, can be written in the classic matrix form:
$$
A^n \mathbf{m}^n = \mathbf{b}^n .
$$
Here, $\mathbf{m}^n$ is the vector of unknown values at the current time step $n$, $A$ is the sparse coefficient matrix, and $\mathbf{b}$ is the vector of known values calculated from the previous time step $n-1$ and anothers.

\paragraph{Coefficient Definitions}
Assuming an $W \cdot H \times W \cdot H$ grid,

The unknown vector $\mathbf{m}^n$ is:
$$
\mathbf{m}^n = \begin{pmatrix} m^n_{1,1} \\ m^n_{1,2} \\ \vdots \\ m^n_{1,H} \\ m^n_{2,1} \\ \vdots \\ m^n_{W,H} \end{pmatrix}, \quad \quad \quad
\textbf{b}^n = \begin{pmatrix} b_{1,1} \\ b_{1,2} \\ \vdots \\ b_{1,H} \\ b_{2,1} \\ \vdots \\ b_{W,H} \end{pmatrix} .
$$

The known vector $\mathbf{b}$ is constructed in the same order, where the entry $b_{i,j}$ corresponding to grid point $(i,j)$ is the entire right-hand side of your equation:
$$
b^n_{i,j} = \frac{m^{n-1}_{i,j}}{\Delta t} - \frac{(f^n_{i+1,j} - f^n_{i-1,j})_x + (f^n_{i,j+1} - f^n_{i,j-1})_y}{2 h} .
$$

\paragraph{The Coefficient Matrix $A$}

The matrix $A$ is a sparse, pentadiagonal matrix where each row $k$ contains the coefficients of the unknown $m^n$ terms for the equation at grid point $(i,j)$.
$$
\mathbf{A} := 
\begin{pmatrix}
\mathbf{D} & \mathbf{U} & 0 & \cdots & 0 \\
\mathbf{L} & \mathbf{D} & \mathbf{U} & \cdots & 0 \\
0 & \mathbf{L} & \ddots & \ddots & \vdots \\
\vdots & & \ddots & \mathbf{D} & \mathbf{U} \\
0 & 0 & \cdots & \mathbf{L} & \mathbf{D}
\end{pmatrix} .
$$

The blocks are \(H \times H\) matrices that describe the connections between grid points.

\paragraph{\(\bullet\) The block \(\mathbf{D}\): A tridiagonal matrix for connections within a grid column}
This block contains the coefficients that link points within the same vertical column (i.e., for a fixed x-index \(i\)).

$$
\mathbf{D} = 
\begin{pmatrix}
C_{\text{diag}} & C_{\text{North}} & 0 & \cdots \\
C_{\text{South}} & C_{\text{diag}} & C_{\text{North}} & \cdots \\
0 & C_{\text{South}} & \ddots & \ddots \\
\vdots & & \ddots & C_{\text{diag}}
\end{pmatrix}_{H \times H} .
$$
where the coefficients are defined as:
\begin{align*}
    C_{\text{diag}} &= \left(\frac{1}{\Delta t} + \frac{2(g^n)^2}{(h)^2}\right), \\
    C_{\text{North}} &= \frac{1}{2h}\left(-\frac{(g^n)^2}{h} + (f^n_{i,j})_y - \frac{1}{4h}(g^n)^2(\tilde{m}^n_{i,j+1} - \tilde{m}^n_{i,j-1})\right), \\
    C_{\text{South}} &= \frac{1}{2h}\left(-\frac{(g^n)^2}{h} - (f^n_{i,j})_y + \frac{1}{4h}(g^n)^2(\tilde{m}^n_{i,j+1} - \tilde{m}^n_{i,j-1})\right).
\end{align*}

\paragraph{\(\bullet\) The blocks \(\mathbf{L}\) and \(\mathbf{U}\): Diagonal matrices for connections between grid columns}
These blocks contain the coefficients linking a point to its neighbors in adjacent columns (East and West).

\begin{itemize}
    \item[\(\mathbf{L}\)] links a column to the \textbf{previous} column (West connections):
    $$ \mathbf{L} = \text{diag}(C_{\text{West}}, C_{\text{West}}, \dots, C_{\text{West}})_{H \times H}, $$
    where
    $$ C_{\text{West}} = \frac{1}{2h}\left(-\frac{(g^n)^2}{h} - (f^n_{i,j})_x + \frac{1}{4h}(g^n)^2(\tilde{m}^n_{i+1,j} - \tilde{m}^n_{i-1,j})\right), $$

    \item[\(\mathbf{U}\)] links a column to the \textbf{next} column (East connections):
    $$ \mathbf{U} = \text{diag}(C_{\text{East}}, C_{\text{East}}, \dots, C_{\text{East}})_{H \times H}, $$
    where
    $$ C_{\text{East}} = \frac{1}{2h}\left(-\frac{(g^n)^2}{h} + (f^n_{i,j})_x - \frac{1}{4h}(g^n)^2(\tilde{m}^n_{i+1,j} - \tilde{m}^n_{i-1,j})\right). $$
\end{itemize}

\subsection{Sample generated by our method}

\begin{figure}[h!]
    \centering
    \includegraphics[width=0.9\linewidth]{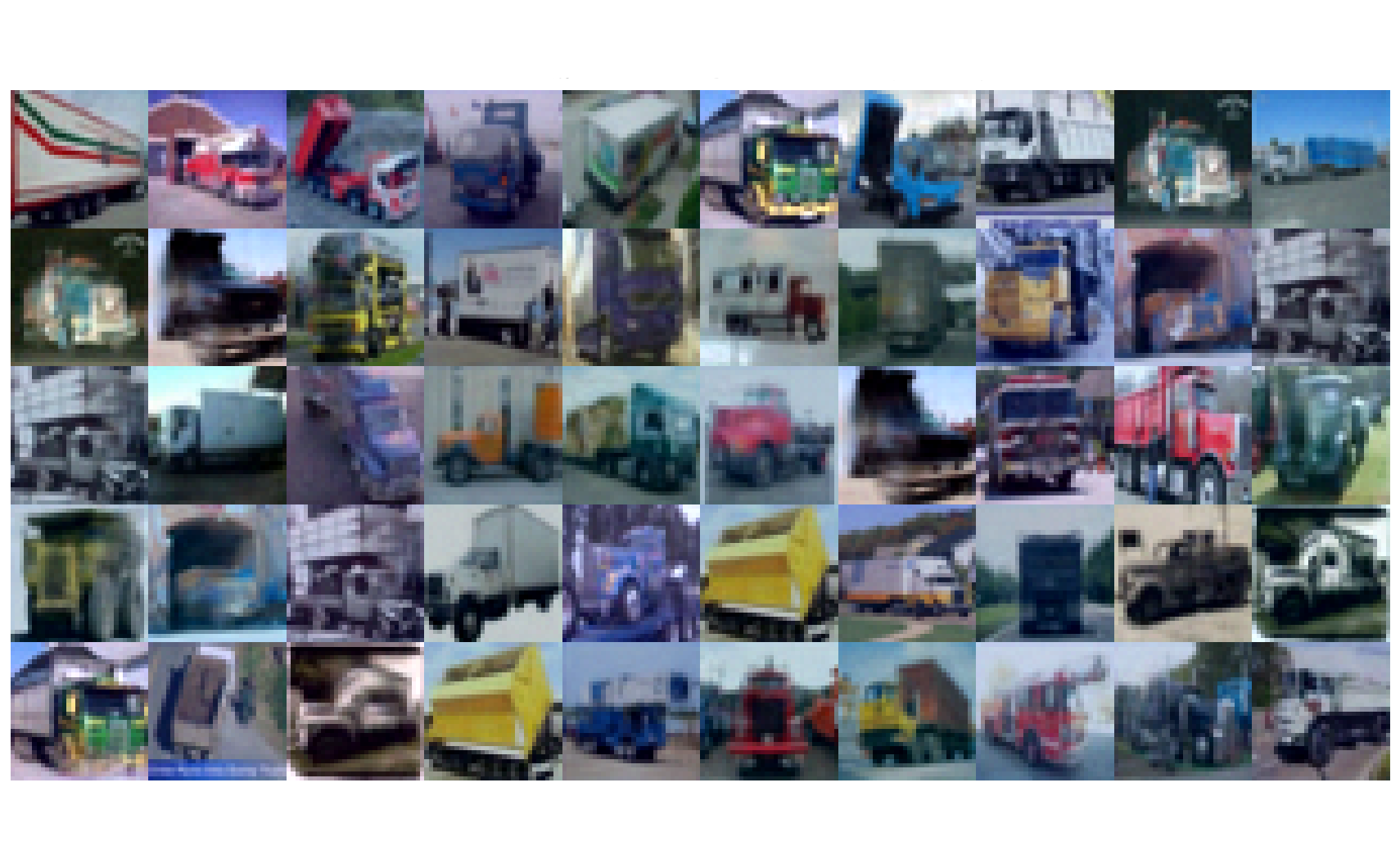}
    \caption{50 32×32 samples generated by our method in experiment 3.}
\end{figure}

\begin{figure}[h!]
    \centering
    \includegraphics[width=0.9\linewidth]{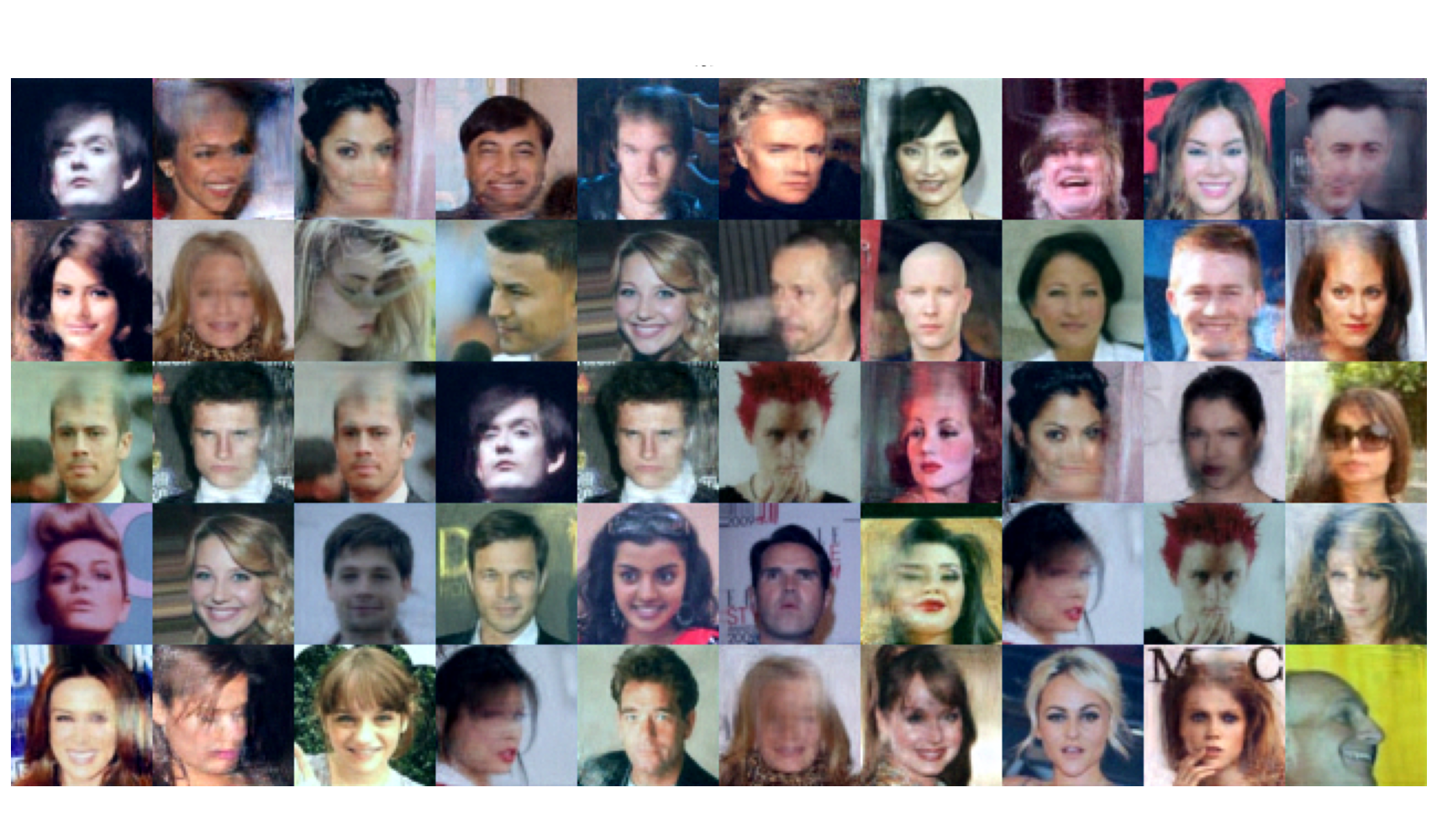}
    \caption{50 64×64 samples generated by our method in experiment 3.}
\end{figure}

\begin{figure}[h!]
    \centering
    \includegraphics[width=0.9\linewidth]{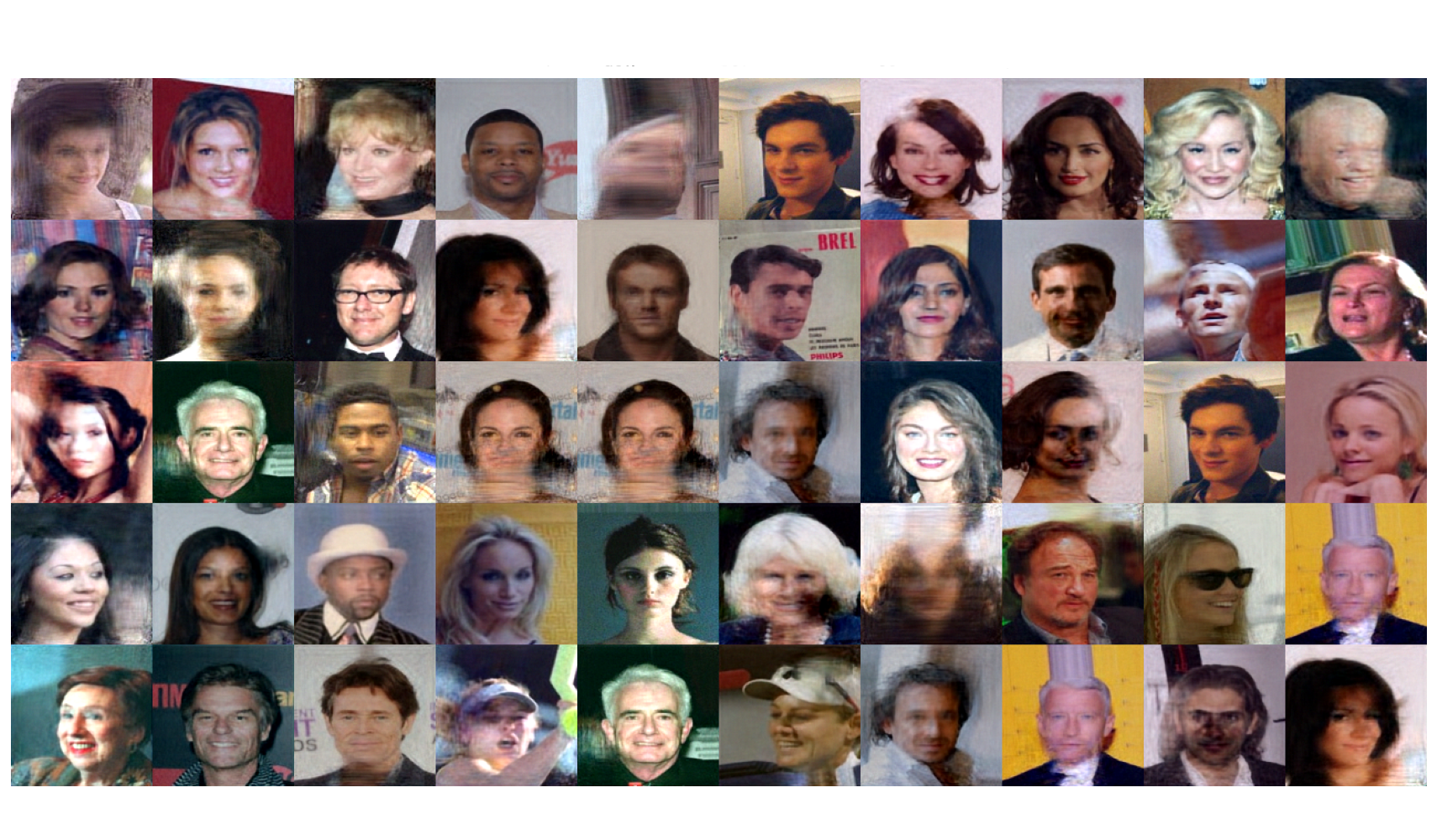}
    \caption{50 128×128 samples generated by our method in experiment 3.}
\end{figure}

\end{document}